\documentclass{article}

\usepackage{microtype}
\usepackage{graphicx}
\usepackage{subcaption}
\usepackage{booktabs} 
\usepackage{tabularray}
\UseTblrLibrary{booktabs}

\usepackage{hyperref}

\usepackage[accepted]{icml2026}

\usepackage{amsmath}
\usepackage{amssymb}
\usepackage{mathtools}
\usepackage{amsthm}

\usepackage{enumitem}
\usepackage{multirow}
\usepackage{graphicx}
\usepackage{wrapfig}
\usepackage{array}
\usepackage{pifont}
\usepackage{bbm}
\usepackage[table,xcdraw]{xcolor}
\usepackage{titletoc} 
\usepackage{algorithm}
\usepackage{algorithmic}
\usepackage{xspace}

\newcommand{\diff}{\text{d}}

\newcommand{\ldriftcoef}{\mathbf{F}_t\mathbf{x}_t}

\newcommand{\ldiffcoef}{\mathbf{G}_t}
\newcommand{\wiener}{\mathbf{w}_t}
\newcommand{\rwiener}{\overline{\mathbf{w}}_t}
\newcommand{\score}{\nabla_{\mathbf{x}_t}{\log{p(\mathbf{x}_t)}}}
\newcommand{\scorec}{\nabla_{\mathbf{x}_t}{\log{p(\mathbf{x}_t\mid\mathbf{x}_0)}}}

\newcommand{\ie}{\emph{i.e.}}
\newcommand{\eg}{\emph{e.g.}}
 
\newcommand{\nall}{S(H)NAP\xspace} 
\newcommand{\nshap}{SHNAP\xspace} 
\newcommand{\gnshap}{gSHNAP\xspace}
 
\newcommand{\ninject}{SNAP\xspace}

\newcommand{\lmpi}{\ensuremath{\text{LMPI}}\xspace}

\newcommand{\her}{\text{RNC}\xspace}

\usepackage[capitalize,noabbrev]{cleveref}

\theoremstyle{plain}
\newtheorem{theorem}{Theorem}[section]

\theoremstyle{definition}

\theoremstyle{remark}

\newtheorem{hypothesis}{Hypothesis}

\usepackage[disable,textsize=tiny]{todonotes}

\icmltitlerunning{Auditing Sybil}

\begin{document}

\twocolumn[
  \icmltitle{Auditing Sybil: Explaining Deep Lung Cancer Risk\\ Prediction Through Generative Interventional Attributions}



  \icmlsetsymbol{equal}{*}

  \begin{icmlauthorlist}
    \icmlauthor{Bartlomiej Sobieski}{equal,uw,wut,ccai}
    \icmlauthor{Jakub Grzywaczewski}{equal,wut,ccai}
    \icmlauthor{Karol Dobiczek}{equal,uj}
    \icmlauthor{Mateusz Wójcik}{wut}
    \icmlauthor{Tomasz Bartczak}{comp}
    \icmlauthor{Patryk Szatkowski}{wum}
    \icmlauthor{Przemysław Bombiński}{wum}
    \icmlauthor{Matthew Tivnan}{mgh,hms,camca}
    \icmlauthor{Przemyslaw Biecek}{uw,wut,ccai}
  \end{icmlauthorlist}

  \icmlaffiliation{uw}{University of Warsaw}
  \icmlaffiliation{wut}{Warsaw University of Technology}
  \icmlaffiliation{ccai}{Centre for Credible AI}
  \icmlaffiliation{uj}{Jagiellonian University}
  \icmlaffiliation{wum}{Medical University of Warsaw}
  \icmlaffiliation{mgh}{Massachusetts General Hospital}
  \icmlaffiliation{hms}{Harvard Medical School}
  \icmlaffiliation{camca}{Center for Advanced Medical Computing and Analysis}
  \icmlaffiliation{comp}{Google}

  \icmlcorrespondingauthor{Bartlomiej Sobieski}{b.sobieski@uw.edu.pl}


  \vskip 0.3in
]



\printAffiliationsAndNotice{}  

\begin{abstract}
    Lung cancer remains the leading cause of cancer mortality, driving the development of automated screening tools to alleviate radiologist workload. Standing at the frontier of this effort is Sybil, a deep learning model capable of predicting future risk solely from computed tomography (CT) with high precision. However, despite extensive clinical validation, current assessments rely purely on observational metrics. This correlation-based approach overlooks the model's actual reasoning mechanism, necessitating a shift to causal verification to ensure robust decision-making before clinical deployment. We propose \nall, a model-agnostic auditing framework that constructs generative interventional attributions validated by expert radiologists. By leveraging realistic 3D diffusion bridge modeling to systematically modify anatomical features, our approach isolates object-specific causal contributions to the risk score. Providing the first interventional audit of Sybil, we demonstrate that while the model often exhibits behavior akin to an expert radiologist, differentiating malignant pulmonary nodules from benign ones, it suffers from critical failure modes, including dangerous sensitivity to clinically unjustified artifacts and a distinct radial bias.
\end{abstract}
\vspace{-2.5em}
\section{Introduction}
\label{sec:introduction}
Lung cancer remains the leading cause of cancer mortality worldwide \citep{Bray2024GlobalCS}, driving global screening efforts with Low-Dose Computed Tomography (LDCT) \citep{NELSON2018WCLC,Pastorino2019MILD}. To address the bottlenecks of radiologist workload and diagnostic variability, automating LDCT interpretation via AI offers a path toward high-throughput, early detection \citep{Arain2025AIAdvances,MDPI2025AIImaging,Field2025AIUKLS}. Standing at the frontier is Sybil \citep{mikhael2023sybil}, a deep learning model predicting 6-year lung cancer risk solely from a single CT scan. Sybil has achieved significant milestones in clinical translation, demonstrating robustness across diverse populations and settings through extensive retrospective and prospective validation \citep{simon2023role,simon2024significance,yang2025deep,kim2025validation,durney2025ma05,aro2024ma02,pasquinelli2025ma05,krule2025ma05,li2025lung}.

\begin{figure}[h]
    \centering
    \includegraphics[width=1.0\linewidth]{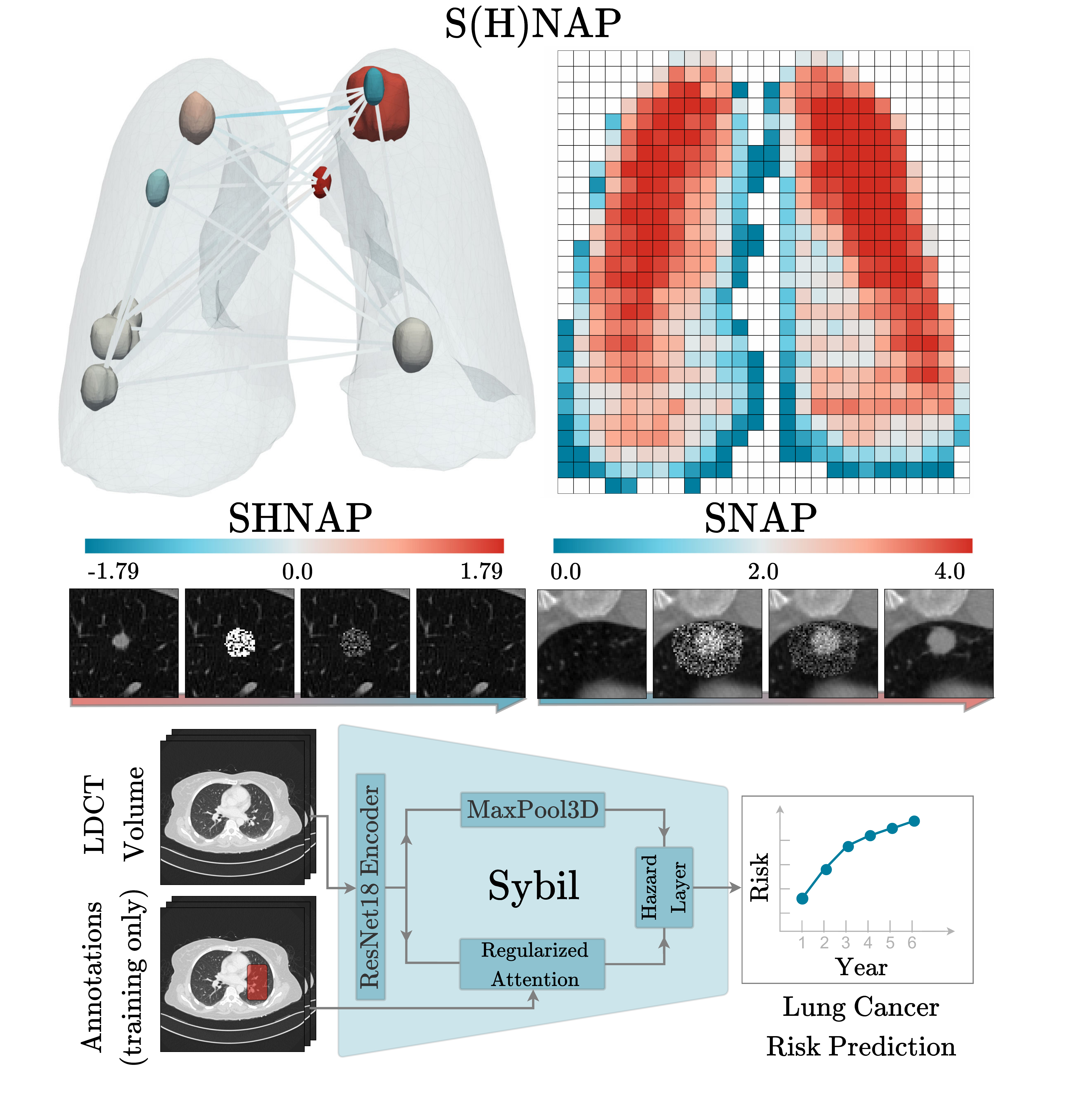}
    \caption{Sybil (\textbf{bottom}) is a frontier model for lung cancer risk prediction from a single CT scan. We propose \nall (\textbf{top}), a novel framework for auditing such models through diffusion-bridge-based generative interventions on pulmonary nodules (\textbf{middle}). \nshap (\textbf{top left}) decomposes the model's prediction into individual nodule contributions and inter-nodule interactions by replacing them with healthy tissue. \ninject (\textbf{top right}) probes volumetric sensitivity by systematically inserting nodules of known malignancy, revealing spatial biases in risk estimation.}
    \vspace{-2.0em}
    \label{fig:teaser}
\end{figure}

However, current assessments rely purely on observational studies. This correlation-based approach confirms \emph{that} the model works, but not \emph{why} or \emph{when} it might fail, creating a critical oversight for high-stakes deployment. To address this, we propose \nall, a framework shifting the paradigm from observational validation to \emph{interventional} auditing. By synergizing game-theoretic attributions, generative modeling, and radiological expertise, we rigorously audit Sybil using a model-agnostic framework applicable to models in lung cancer screening scenarios. Specifically, \textbf{1.} we introduce novel methodologies for synthetic interventions on pulmonary nodules in 3D LDCT using diffusion bridge modeling. \textbf{2.} We construct \nshap and \ninject, intervention-based attribution methods tailored for lung cancer risk prediction. \textbf{3.} We empirically verify that Sybil functions as a linear model with pairwise interactions over pulmonary nodules. \textbf{4.} We reveal critical misalignments, including a radial sensitivity bias and learned spurious artifacts, highlighting the necessity of interventional auditing.

\section{Related works}
\label{sec:related_works}
\textbf{Sybil.} We define Sybil \citep{mikhael2023sybil} as a parameterized distribution $p_{\boldsymbol{\theta}} (\mathbf{y}\mid\cdot)$ over lung cancer risk $\mathbf{y}=(y_i)_{i=0}^{6}\in[0, 1]^7$. The outputs consist of a base risk $y_0$ and cumulative risks $y_i=y_0 + \sum_{j=1}^i O_j$ for $i=1,\dotsc,6$ ($O_i \geq 0$). We restrict our analysis to the \emph{base hazard} logit $f_{\boldsymbol{\theta}} (y_0\mid\cdot)$ (where $p_{\boldsymbol{\theta}} (y_0\mid\cdot)=\sigma(f_{\boldsymbol{\theta}} (y_0\mid\cdot))$), as the base risk correlates almost perfectly with cumulative outputs (see \cref{fig:risk_correlation}) and Sybil's performance decays over time (\citealp{mikhael2023sybil}; fig. 2). Structurally (\cref{fig:teaser}), Sybil employs a 3D ResNet18 encoder \citep{he2016deep} whose features are processed by separate max-pooling and attention branches \citep{vaswani2017attention} to compute final estimates. The attention mechanism is additionally regularized to match expert annotations in the form of segmentation masks. The model was developed using a large-scale cohort of over 28,000 scans from the NLST \citep{NLST2011}.

\textbf{Attribution methods.} Most common explanation methods for deep neural networks assign scalar importance scores to input features. Various approaches explored diverse importance definitions and constraints \citep{simonyan2013deep,bach2015pixel,shrikumar2017learning,sundararajan2017axiomatic,selvaraju2020grad}. Grounded in game theory \citep{shapley1953value}, SHapley Additive exPlanations (SHAP; \citealp{lundberg_shap}) are a specific instance of \emph{additive} feature attribution methods. SHAP uniquely satisfies the interpretablity axioms of \emph{local accuracy}, \emph{missingness}, and \emph{consistency}. Recent work emphasizes evaluating models on in-distribution inputs, improving saliency maps by respecting the data manifold \citep{zahermanifold,ademi2025pomelo,salek2025using}. In \cref{app:xai_medical}, we provide a broader overview of XAI techniques applied to medical imaging models.

\textbf{Counterfactual explanations.} Occupying the highest rung of Pearl's causality ladder \citep{Pearl_2009}, visual counterfactual explanations (VCEs) aim to modify a sample in a minimal and semantically meaningful way to alter a model's prediction. Unlike standard attributions, VCEs enable \emph{cause-and-effect} analysis of \emph{what-if} scenarios. To ensure modifications remain on the data manifold, most approaches leverage generative models to approximate the underlying distribution \citep{jacob2022steex,jeanneret2022diffusion,augustin2022diffusion,jeanneret2023adversarial,augustin2024digin,jeanneret2024text,sobieski2024global,sobieski2025rethinking}. Most relevant to this work are \emph{region-constrained} VCEs \citep{sobieski2025rethinking}, which employ guided image-to-image diffusion bridges \citep{liu20232,sobieski2025sdb} to localize edits in natural images.

\section{Background}
\label{sec:background}

\textbf{Notation.} Let $[d]=\{1,\dots,d\}$ denote feature indices. For any subset $S\subseteq[d]$, we define $\mathbf{x}_S$ as the feature sub-vector restricted to indices in $S$. Similarly, $f_S(\mathbf{x}_S)$ denotes a component function depending exclusively on $S$.

\textbf{Linear Models with Pairwise Interactions.} Linear models are widely recognized as inherently \emph{interpretable} tools for explaining data mechanisms \citep{hastie2009elements, rudin2019stop}. To enlarge their scope while retaining this \emph{white-box} nature, they are often extended to include \emph{pairwise interactions}. We term $f:\mathbb{R}^d\rightarrow\mathbb{R}$ a \emph{linear model with pairwise interactions} (\lmpi) if it decomposes as:
\begin{equation}\label{eq:lmpi}
    f(\mathbf{x}) = \underbrace{\beta_0}_{\text{Intercept}} + 
    \underbrace{\sum_i \beta_i x_i}_{\text{Main effects}} + 
    \underbrace{\sum_{1\leq i<j \leq d}\beta_{ij}x_i x_j}_{\text{Pairwise interactions}},
\end{equation}
where $\beta_0, \beta_i, \beta_{i,j} \in \mathbb{R}$ are learnable coefficients.

\textbf{n-Shapley Values.} Shapley-based Interaction Indices (SII, \citealp{grabisch1999axiomatic}) extend additive attributions to capture feature dependencies \citep{sundararajan2020shapley,tsai2023faith}. For a model $f$ and sample $\mathbf{x}$, let the \emph{value function} $v_{\mathbf{x}} (T)$ represent $f$'s predictions using only features $T\subseteq[d]$. Its \emph{discrete derivative} $\Delta_{S}v_{\mathbf{x}} (T)=\sum_{L\subseteq S} (-1)^{|S|-|L|}v_{\mathbf{x}} (T\cup L)$ isolates the pure interaction effect of $S$ given $T$.

To explain $f(\mathbf{x})$, \emph{n-Shapley Values} (nSV, \citealp{bordt2023shapley}) approximate the prediction via:
\begin{align}\label{eq:n_shapley_values}
    f(\mathbf{x}) &\approx \sum_{S\subseteq [d], 0 \leq |S| \leq n} \phi_S(\mathbf{x}), \text{ where}\quad \\
    \phi_S(\mathbf{x}) &= \sum_{T \subseteq [d] \setminus S} \frac{(d - |T| - |S|)! \, |T|!}{(d - |S| + 1)!} \Delta_S v_{\mathbf{x}} (T).
\end{align}
For brevity, we omit the dependence of $\phi$ on $\mathbf{x}$ (denoting $\phi_S$) in the remainder of the text.

In this work, we use $n=2$ with the \emph{interventional} value function $v_{\mathbf{x}} (S)=\mathbb{E}_{\mathbf{z}\sim \mathcal{D}}[f(\mathbf{x}_{S}, \mathbf{z}_{[d]\setminus S})]$, which recovers the structure of a pairwise interaction model, where $\phi_\varnothing$ corresponds to the baseline value. Notably, nSV constitutes the unique least-squares projection of the set function $v_{\mathbf{x}}$ onto the space of additive games of order $n$, effectively approximating the local decision boundary as an \lmpi.

To measure the quality of this approximation, we use the coefficient of determination over the feature lattice:
\begin{equation}\label{eq:r2}
R^2 = 1 - \frac{\sum_{S \subseteq [d]} \left( v_{\mathbf{x}} (S) - \hat{v}_{\text{nSV}} (S) \right)^2}{\sum_{S \subseteq [d]} \left( v_{\mathbf{x}} (S) - \bar{v}_{\mathbf{x}} \right)^2}, 
\end{equation}
where $\bar{v}_{\mathbf{x}}=2^{-d}\sum_{S\subseteq[d]}v_{\mathbf{x}} (S)$ is the mean value and $\hat{v}_{\text{nSV}} (S) = \sum_{K \subseteq S, |K| \le n} \phi_K(\mathbf{x})$ is the value predicted by the nSV approximation for subset $S$. In XAI, \cref{eq:r2} is termed \emph{unweighted local fidelity} \citep{garreau2020explaining}.

\textbf{Bridging with diffusion.} Generative modeling in computer vision is currently dominated by \emph{diffusion models} (DMs, \citealp{ho2020denoising,songscore}). Formally, DMs define a \emph{forward} process (\cref{eq:forward_diffusion}) mapping data $p(\mathbf{x})$ to a latent state over time $t\in[0,1]$, and a \emph{reverse} generative process (\cref{eq:reverse_diffusion}), governed by the SDEs:
\begin{align}
    \diff \mathbf{x}_t &= \ldriftcoef \diff t + \ldiffcoef \diff \wiener,\label{eq:forward_diffusion}\\ 
    \diff \mathbf{x}_t &= [\ldriftcoef - \ldiffcoef \ldiffcoef^\top \score] \diff t + \ldiffcoef \diff \rwiener, \label{eq:reverse_diffusion}
\end{align}
where $\mathbf{F}_t, \ldiffcoef$ are the drift and diffusion coefficients, and the \emph{score} $\score$, approximated by the neural network $\mathbf{s}_{\boldsymbol{\xi}}$, drives generation through numerical integration. While standard DMs map to noise, \emph{System-Embedded Diffusion Bridges} (SDB, \citealp{sobieski2025sdb}) generalize the endpoint to a linear measurement $\mathbf{x}'=\mathbf{A}\mathbf{x} + \boldsymbol{\Sigma}^{\frac{1}{2}}\boldsymbol{\varepsilon}$ at $t=1$. When $\mathbf{A}$ is a binary mask and $\boldsymbol{\Sigma}=\mathbf{0}$, SDB functions as a specialized inpainting model, initializing from masked input $\mathbf{x}'$ to restore missing content. Crucially, this confines the diffusion strictly to unobserved regions, leaving known content unperturbed.

\section{Methodology}
\label{sec:methodology}

\textbf{Motivation.} Current attribution methods face a trade-off: either highlight individual feature importance while staying on the data manifold \citep{zahermanifold,ademi2025pomelo,salek2025using} or measure feature interactions while evaluating in ambient space \citep{sundararajan2020shapley,tsai2023faith,bordt2023shapley}. Conversely, VCEs provide in-distribution causal explanations but fail to isolate specific feature contributions. To capture both feature importance and mutual influence while adhering to the data manifold, we propose novel attribution methods based on generative interventions. These leverage domain knowledge to simplify the feature space, making them specifically suited for lung cancer risk prediction.

Grounded in the clinical consensus that pulmonary nodules are the primary predictive biomarkers for lung cancer \citep{callister2015british,macmahon2017guidelines,wood2025nccn}, we formulate the following hypothesis:
\begin{hypothesis}\label{hyp:additive_decomposition} For a given sample, Sybil's decision function can be effectively approximated by an \lmpi consisting of a sample-specific background term and a model of main effects and pairwise interactions over pulmonary nodules.
\end{hypothesis}
Formally, let $\mathbf{x}$ represent a given patient's CT scan. \Cref{hyp:additive_decomposition} states that 
\begin{equation}\label{eq:shnap}
    f_{\boldsymbol{\theta}} (y_0\mid\mathbf{x}) \approx \mu_{\mathbf{x}} + \sum_{i=1}^N \phi_i n_i + \sum_{i=1}^N \sum_{j > i} \phi_{ij} n_i n_j,
\end{equation} 
where $n_i \in \{0,1\}$ indicates the presence of the $i$-th nodule, $\phi$ coefficients capture main and pairwise effects, $N$ is the nodule count, and $\mu_{\mathbf{x}}$ represents the sample-specific baseline (nodule-free scan). Notably, for a given $\mathbf{x}$, \cref{eq:n_shapley_values} constitutes the unique least-squares solution to \cref{eq:lmpi}, thereby fully determining the coefficients for \cref{eq:shnap}.

Validity of \cref{eq:shnap} implies Sybil processes nodules as modular semantic units, independent of the background context. Consequently, introducing a nodule should shift predictions in a structurally predictable manner governed by its main and interaction effects. This modularity facilitates nodule-centric counterfactuals, isolating causal impacts without confounding effects from the anatomical background.

\textbf{Methodological Gap.} The primary challenge in estimating \cref{eq:shnap} is the lack of paired counterfactual data, \ie, scans where nodule coalitions are selectively modified. To bridge this, we leverage SDB as a generative proxy for the LDCT distribution. We design a framework utilizing this prior to perform reliable semantic interventions, enabling both the replacement of nodules with healthy tissue and the synthesis of realistic nodules with controlled properties.

\subsection{Synthetic perturbations}
\begin{figure*}[htbp]
    \centering
    \includegraphics[width=0.98\linewidth]{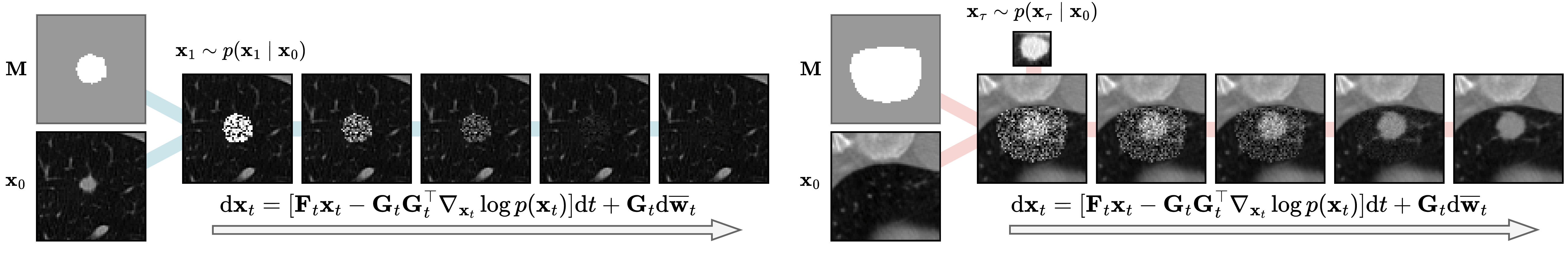}
    \caption{2D visualizations of our nodule removal (\textbf{left}) and insertion (\textbf{right}) approaches performed on 3D subvolumes of an LDCT scan.}
    \vspace{-1em}
    \label{fig:nshap_ninject}
\end{figure*}

\textbf{Distribution blending.} To unify SDB with nodule removal and insertion, we invoke the theoretical result of \citet{verdu2009mismatched} on how distributions \emph{blend} under diffusion.

\begin{theorem}[Mismatched estimation, \citealp{verdu2009mismatched}]\label{th:distribution_blending} 
Let $p(\mathbf{x}^1)$ and $p(\mathbf{x}^2)$ be two probability distributions with time-parameterized evolutions $p(\mathbf{x}_t^1)$ and $p(\mathbf{x}_t^2)$ under the forward process (\cref{eq:forward_diffusion}). The Kullback-Leibler divergence between them decomposes as:
\begin{equation}
\begin{split}
    D_{KL} (p(\mathbf{x}_t^1) \| p(\mathbf{x}_t^2)) &= D_{KL} (p(\mathbf{x}_0^1) \| p(\mathbf{x}_0^2)) \\
    &\quad - \frac{1}{2} \int_0^t \mathcal{J}_{\mathbf{D}_\tau} (\tau) \, d\tau,
\end{split}
\end{equation}
where $\mathbf{D}_\tau = \mathbf{G}_\tau \mathbf{G}_\tau^T$ and $\mathcal{J}$ is the Relative Fisher Information (RFI):
\begin{equation}
    \mathcal{J}_{\mathbf{D}_\tau} (\tau) = \mathbb{E}_{\mathbf{x} \sim p(\mathbf{x}_\tau^1)} \Bigl[ \Bigl\| \nabla_{\mathbf{x}} \log p(\mathbf{x}_\tau^1) - \nabla_{\mathbf{x}} \log p(\mathbf{x}_\tau^2) \Bigr\|_{\mathbf{D}_\tau}^2 \Bigr].
\end{equation}
\end{theorem}
\vspace{-1em}
Crucially, \cref{th:distribution_blending} states that $p(\mathbf{x}_t^1)$ and $p(\mathbf{x}_t^2)$ become increasingly indistinguishable over time due to the non-negativity of RFI. Consequently, for a score model trained on $p(\mathbf{x}^1)$, diffused samples from a different distribution $p(\mathbf{x}^2)$ become statistically indistinguishable from $p(\mathbf{x}_t^1)$ after some timestep. While implicitly relied upon in image editing \citep{choi2021ilvr,meng2022sdedit,lugmayr2022repaint,su2023dual,couairon2023diffedit} and seemingly connected to recent fluctuation theory \citep{ramachandran2025crossfluctuation}, this theoretical justification is rarely explicitly formulated. Here, we leverage the general matrix-valued SDE to link this result directly to SDB.

\textbf{Nodule removal.} Let $p(\mathbf{x}^1)$ be the LDCT training distribution for $\mathbf{s}_{\boldsymbol{\xi}}$ and $p(\mathbf{x}^2)$ be a counterfactual distribution where a specific nodule (within mask $\mathbf{A}$) is replaced with healthy tissue via a \emph{do}-operator \citep{Pearl_2009}. Applying \cref{th:distribution_blending} to their forward evolution up to $t=1$, the distinguishing information within $\mathbf{A}$ vanishes. SDB guarantees this erasure, as its dynamics converge to a state defined entirely by the masked region at $t=1$. Since pulmonary nodules rarely exceed 0.1\% of total lung volume \citep{horeweg2014lung}, the score model $\mathbf{s}_{\boldsymbol{\xi}}$ effectively acts as a healthy tissue prior during reverse sampling, justifying this procedure for in-distribution nodule removal.

\textbf{Nodule insertion.} As anatomical anomalies linked to malignancy, nodules are unlikely to be generated by an unconditional model. To insert them, we define $p(\mathbf{x}^2)$ as a copy of $p(\mathbf{x}^1)$ where a nodule from a different patient is transplanted into mask $\mathbf{A}$ and aligned with the new context. \Cref{th:distribution_blending} implies $p(\mathbf{x}_t^1)$ and $p(\mathbf{x}_t^2)$ become statistically indistinguishable at some specific timestep $t=\tau$. In practice, we simulate $p(\mathbf{x}^2)$ by "copy-pasting" a specific nodule prior to diffusion. The theorem guarantees that at $t=\tau$, the trained score model treats the diffused state as a valid sample from $p(\mathbf{x}_\tau^1)$, allowing the reverse process to coherently integrate the inserted nodule into the surrounding anatomy.

Examples of both nodule removal and insertion procedures are visualized in \cref{fig:nshap_ninject}, while broader literature context is provided in \cref{app:removal_insertion}.

\subsection{Constructing the explanations}

\textbf{Explaining by removing.} Nodule removal enables the \lmpi approximation in \cref{eq:shnap}. Let $\mathcal{N}=\{1,\dots,N\}$ denote the set of detected nodules. For any subset $S \subseteq \mathcal{N}$, let $\mathbf{x}_S$ be the scan where nodules in $S$ are preserved and those in $\mathcal{N} \setminus S$ are replaced with healthy tissue. We generate the collection of samples $\mathcal{X} = \{ \mathbf{x}_S : S \subseteq \mathcal{N} \}$ covering all possible coalitions of nodules. We pair these with Sybil's responses to form the dataset $D=\{ (S, v_{\mathbf{x}} (S)) \}_{\mathbf{x}_S \in \mathcal{X}}$, where $v_{\mathbf{x}} (S) = f(y_0\mid\mathbf{x}_S)$. This requires only $N$ SDB removal trajectories to construct the base $\mathbf{x}_\varnothing$ and components for linear recomposition, and $2^N$ evaluations of Sybil, which is computationally inexpensive for a realistic number of lung nodules. We then solve for the coefficients using nSV\footnote{Extending the SHAP-IQ package \citep{Muschalik.2024b}.} due to its uniqueness and axiomatic properties (\citealp{lundberg_shap,garreau2020explaining}; \cref{app:shap_nsv}). We term this procedure \textbf{\emph{SHapley Nodule Attribution Profiles} (\nshap)}. For pseudocode, see \cref{alg:shnap}.

\textbf{Explaining by inserting.} Inserting nodules with known properties allows verifying Sybil's reasoning and spatial sensitivity. Given a nodule volume extracted from a patient using a mask $\mathbf{A}$, let $\mathbf{r}$ denote the nodule's volumetric content. We align this content to be centered at a target coordinate $\mathbf{c}=(i, j, k)$ in a different scan $\mathbf{x}$ using our insertion protocol. We assign an attribution score based on the resulting prediction shift:
\begin{equation}\label{eq:snap}
    \psi_{\mathbf{c}} = f(y_0\mid \mathbf{x}_{\mathbf{c} \leftarrow \mathbf{r}}) - f(y_0 \mid \mathbf{x}),
\end{equation}
where $\mathbf{x}_{\mathbf{c} \leftarrow \mathbf{r}}$ denotes the scan with the nodule $\mathbf{r}$ inserted at location $\mathbf{c}$. As $f$ represents the logit, \cref{eq:snap} corresponds to the log-odds ratio between the intervened and initial states. We term this procedure \textbf{\emph{Substitutive Nodule Attribution Probing} (\ninject)}. For pseudocode, see \cref{alg:snap}.
\vspace{-1em}
\section{Experiments}
\label{sec:experiments}
\textbf{Datasets \& Implementation.} We utilize three 3D LDCT datasets: \textbf{D1.} NLST \citep{NLST2011} (approx. 28,000 training and 6,000 test scans) for SDB training; \textbf{D2.} LUNA25 \citep{Peeters2025LUNA} (4,069 scans), featuring biopsy-confirmed malignancy or 2-year stability verification for benign cases, for which we construct naive spherical masks from nodule coordinates; and \textbf{D3.} iLDCT (243 scans), an internal out-of-distribution testbed with a higher prevalence of severe cases and precise expert radiologist annotations. Lung and lobe segmentations are obtained using the improved TotalSegmentator \citep{wasserthal2023totalsegmentator} proposed by \citet{10943755}. To manage computational constraints, we train a discrete-time (1000 steps) Schrödinger Bridge (SB) variant of SDB on randomly sampled $64^3$ cubes, generating training masks procedurally via metaballs \citep{blinn1982generalization}. We compare SDB reconstruction performance with other baselines for CT synthesis. Both nodule removal and insertion are performed with 100 NFE. For details, see \cref{app:experimental_setup}.
\vspace{-1em}
\subsection{Validity of synthetic perturbations}\label{subsec:expert_study}
\begin{figure}[htbp]
    \centering
    \includegraphics[width=0.86\linewidth]{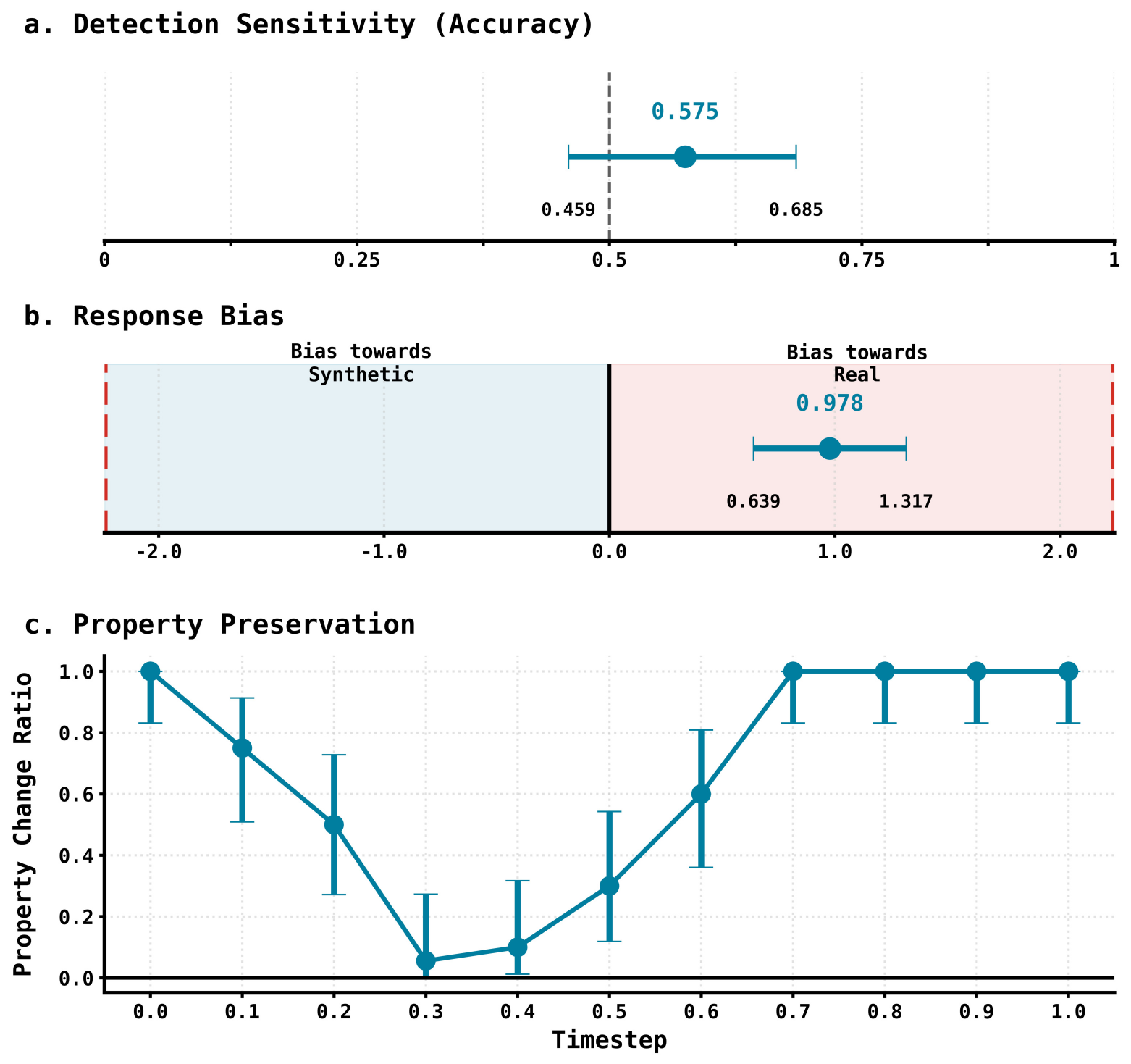}
    \caption{Results of an expert study evaluating the realism of nodule removal (\textbf{a.}, accuracy; \textbf{b.}, response bias) and the preservation of properties during nodule insertion (\textbf{c.}).}
    \vspace{-1em}
    \label{fig:expert_study}
\end{figure}
To ensure our perturbations remain in-distribution, a prerequisite for valid explanations, we conducted a study with two board-certified radiologists.

\textbf{Nodule removal.} We performed a blinded evaluation where radiologists assessed 40 3D cubes from NLST (20 real healthy tissue, 20 synthetic removal) in a binary classification task distinguishing real tissue from synthetic. \Cref{fig:expert_study} (\textbf{a}) shows that their performance is statistically indistinguishable from random guessing (exact binomial test, point estimate 0.57). Furthermore, evaluation of response bias ($c$, \citealp{Macmillan2005}) reveals a significant tendency to label samples as ``real'' (\Cref{fig:expert_study} (\textbf{b}), single-sample Z-test), confirming the high fidelity of the generations.

\textbf{Nodule insertion.} The parameter $\tau$ controls the trade-off between preserving source content and aligning with the new context. Radiologists evaluated 110 pairs of scans (original vs. inserted) across $\tau \in \{0, 0.1, \dots, 1.0\}$. For each pair, they assessed whether the synthetic nodule exhibited perceptible deviations from the reference regarding structural properties, malignancy characteristics, or background alignment. \Cref{fig:expert_study} (\textbf{c}) presents 95\% CIs from an exact binomial test for each $\tau$, illustrating the transition from artifact-heavy naive insertion ($\tau \to 0$) to excessive deviation ($\tau \to 1$). The optimal balance is achieved at $\tau=0.3$, where nodules consistently preserve source properties without visual artifacts. We use this value for all subsequent experiments.
\subsection{Validity of the explanations}\label{subsec:validity_additive}
\begin{figure}[htbp]
    \centering
    \includegraphics[width=0.9\linewidth]{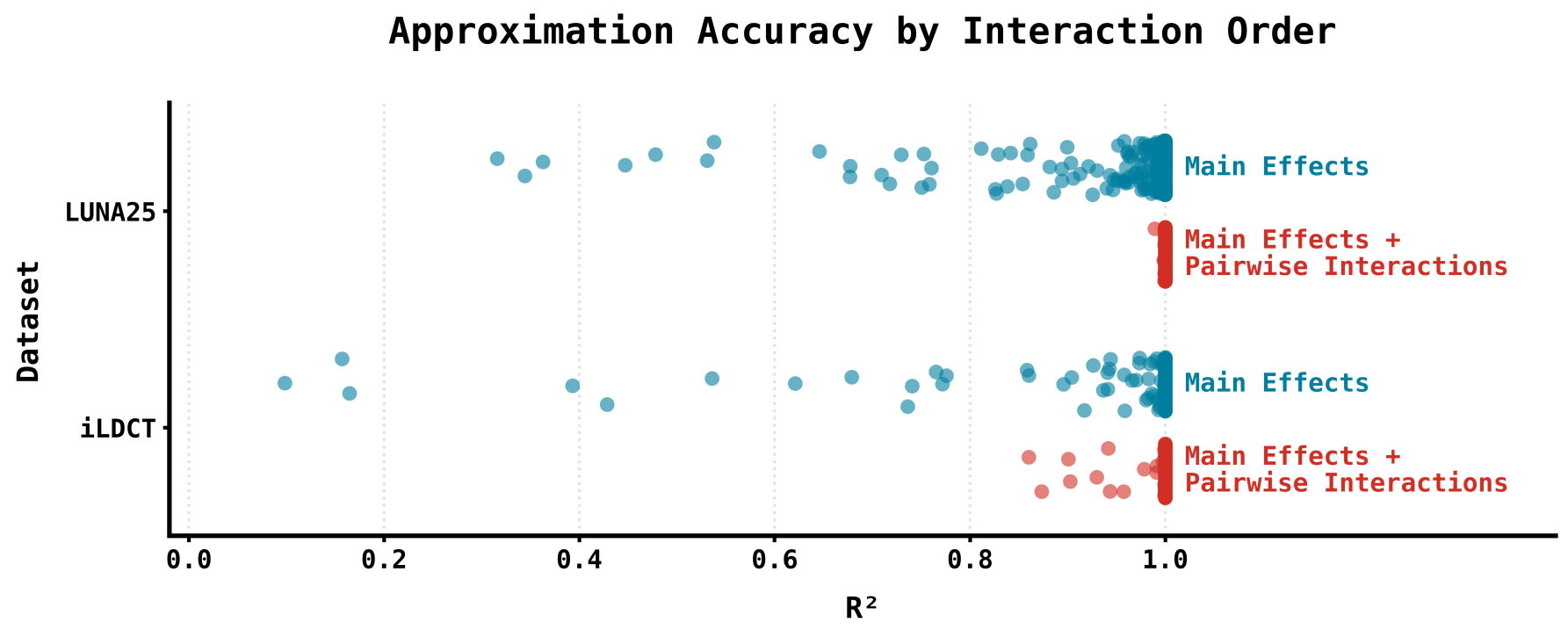}
    \caption{Accuracy of approximating Sybil as an \lmpi over pulmonary nodules across two datasets, including both first- and second-order effects.}
    \vspace{-1em}
    \label{fig:validity_additive}
\end{figure}

\textbf{Approximation accuracy.} Establishing that nodule removal results in indistinguishable healthy tissue confirms that \nshap explanations stem from in-distribution perturbations. This allows verifying \cref{hyp:additive_decomposition}: that Sybil is effectively an \lmpi. We compute \nshap on the entire LUNA25 test split and iLDCT, evaluating the fit via $R^2$ (\cref{eq:r2}). \Cref{fig:validity_additive} shows results for main effects and pairwise interactions. Main effects alone suffice for a perfect fit in the majority of cases, evidenced by the collapsed interquartile range at $R^2 \approx 1$. The outlier tail is almost entirely eliminated by adding pairwise interactions. Remaining failures typically correspond to rare, anomalously large nodules where SDB struggles with reconstruction; a limitation addressable by training on larger volumes. Overall, these results strongly support \cref{hyp:additive_decomposition}.

\textbf{Counterfactual tractability.} While \cref{fig:validity_additive} validates \nshap, it also reinforces the reliability of \ninject. Confirming that Sybil functions as an \lmpi implies it processes nodules additively. Consequently, synthetic insertion adheres to this same mechanism, ensuring mathematically consistent counterfactuals and transparent attributions.

\textbf{Stability.} To evaluate robustness, \cref{fig:validity_stability} displays the density of standard deviations for \nshap attributions across 5 independent runs, where stochasticity arises from the random initial state of the reverse diffusion process. The values concentrate heavily around zero, indicating a systematic response to healthy tissue imputation and minimizing the risk of adversarial artifacts. We also compare \nshap to naive perturbations (\cref{app:naive_stability}); the latter reveals that out-of-distribution inputs result in unstable attributions, highlighting the necessity of generative interventions.

\subsection{Opening Sybil's black box with \nshap}

Building on the foundation that Sybil can be effectively represented as a \lmpi over pulmonary nodules, we proceed to use \nshap to dissect its decision-making mechanisms.

\begin{figure}[htbp]
    \centering
    \includegraphics[width=0.45\linewidth]{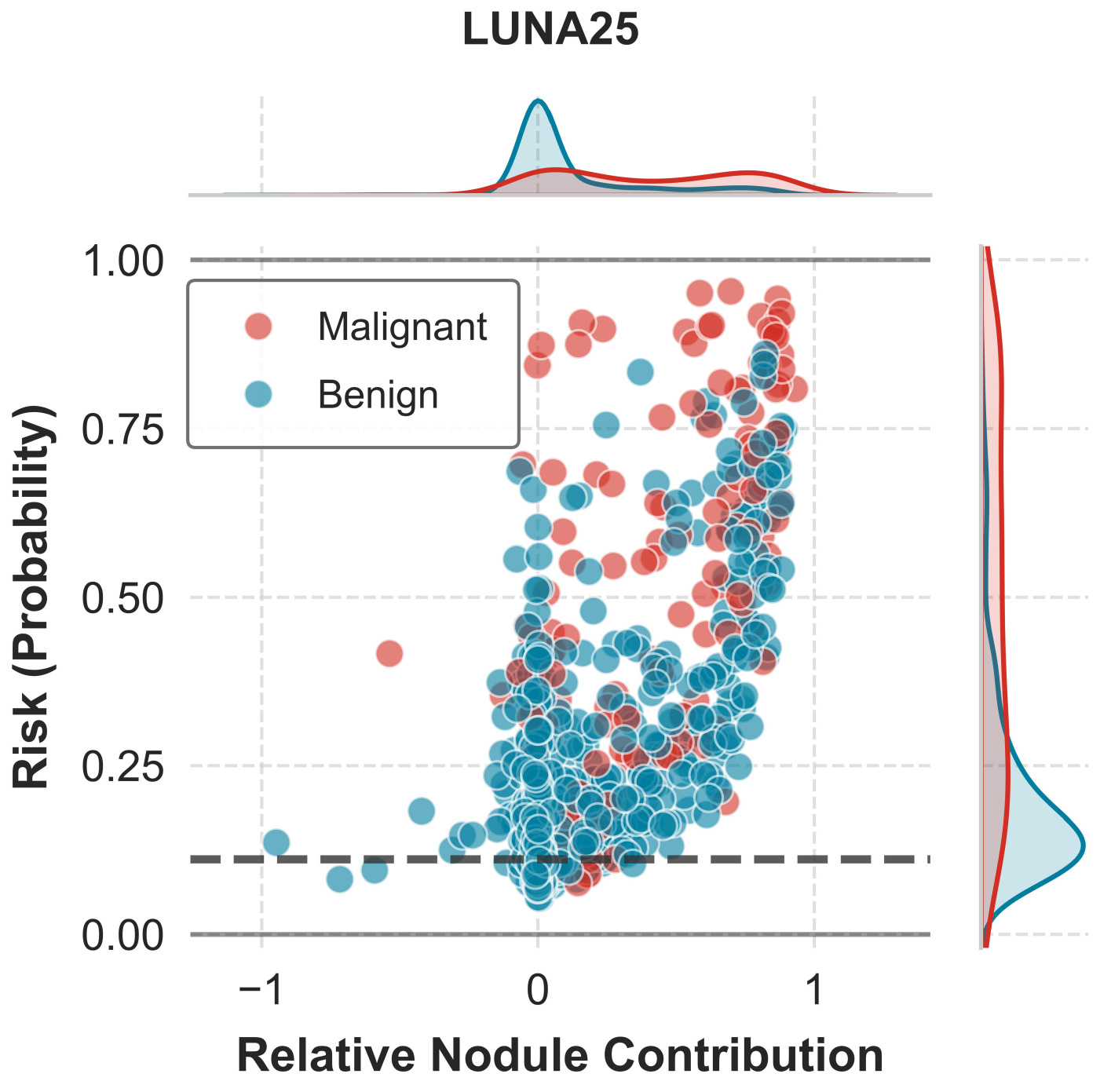}
    \includegraphics[width=0.45\linewidth]{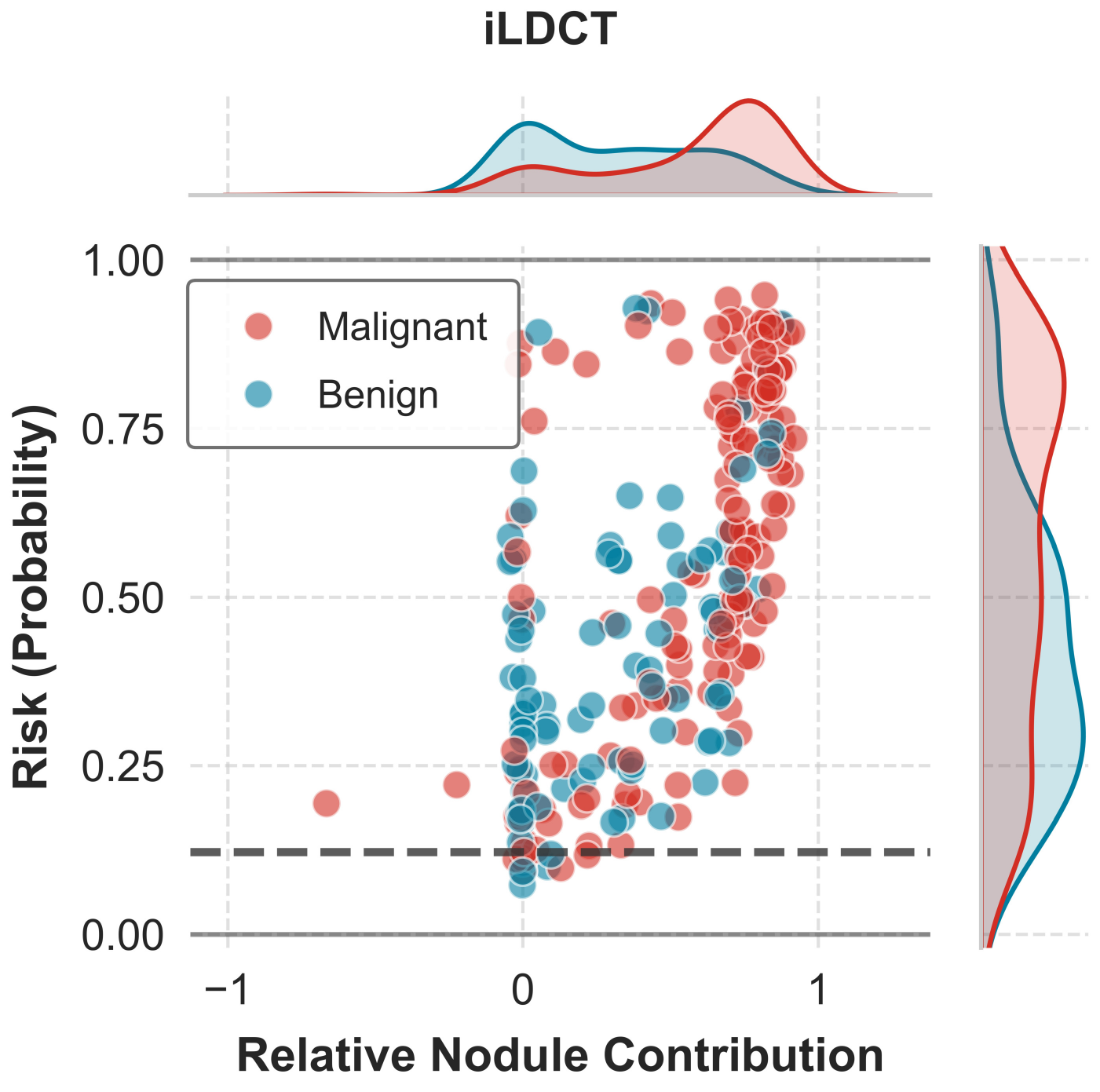}
    \caption{Comparison of risk predicted by Sybil and relative nodule contributions across two datasets.}
    \label{fig:nshap_statistics}
\end{figure}
\vspace{-1em}
\textbf{When pulmonary nodules matter most.} A unique advantage of our decomposition (\cref{eq:shnap}) is the separation of nodule-specific effects from the anatomical background $\mu_\mathbf{x}$. We leverage this to move beyond manual auditing, defining the \emph{Relative Nodule Contribution} (\her) as $\her(\mathbf{x}) = \frac{\sigma(f(y_0\mid\mathbf{x})) - \sigma(\mu_{\mathbf{x}})}{\sigma(f(y_0\mid\mathbf{x}))}$ to detect cases where model reasoning may be flawed. Here, $\sigma(\cdot)$ converts the logits $f$ to probabilities. To interpret risk levels, we set a decision threshold on the test split enforcing $\geq 95\%$ sensitivity, reflecting the clinical priority of minimizing false negatives.

\begin{figure*}[htbp] 
    \centering
    \vspace{-3.5em} 
    \setkeys{Gin}{width=0.32\textwidth, trim=0 20 0 20, clip} 
    
    \includegraphics{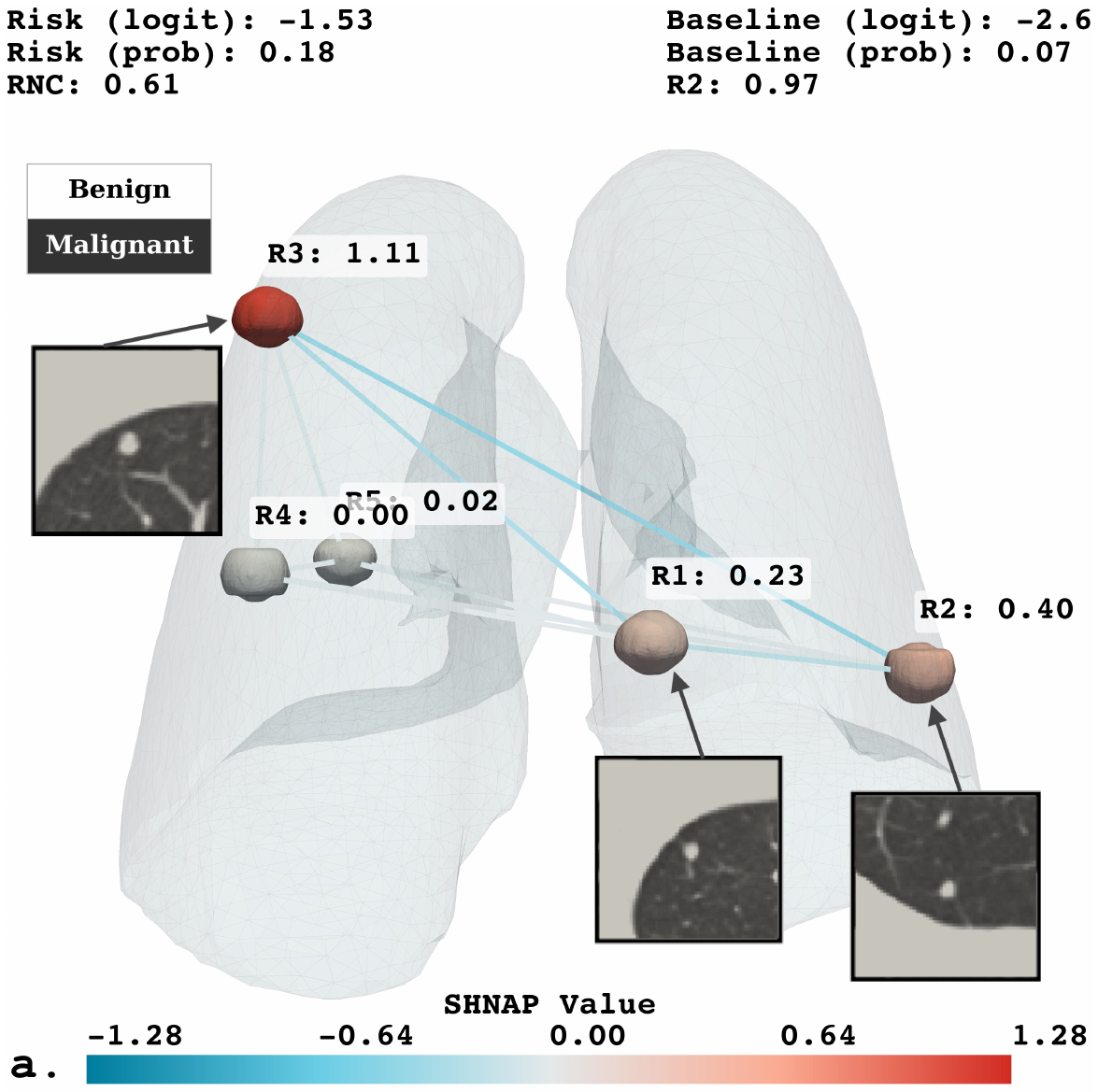}
    \includegraphics{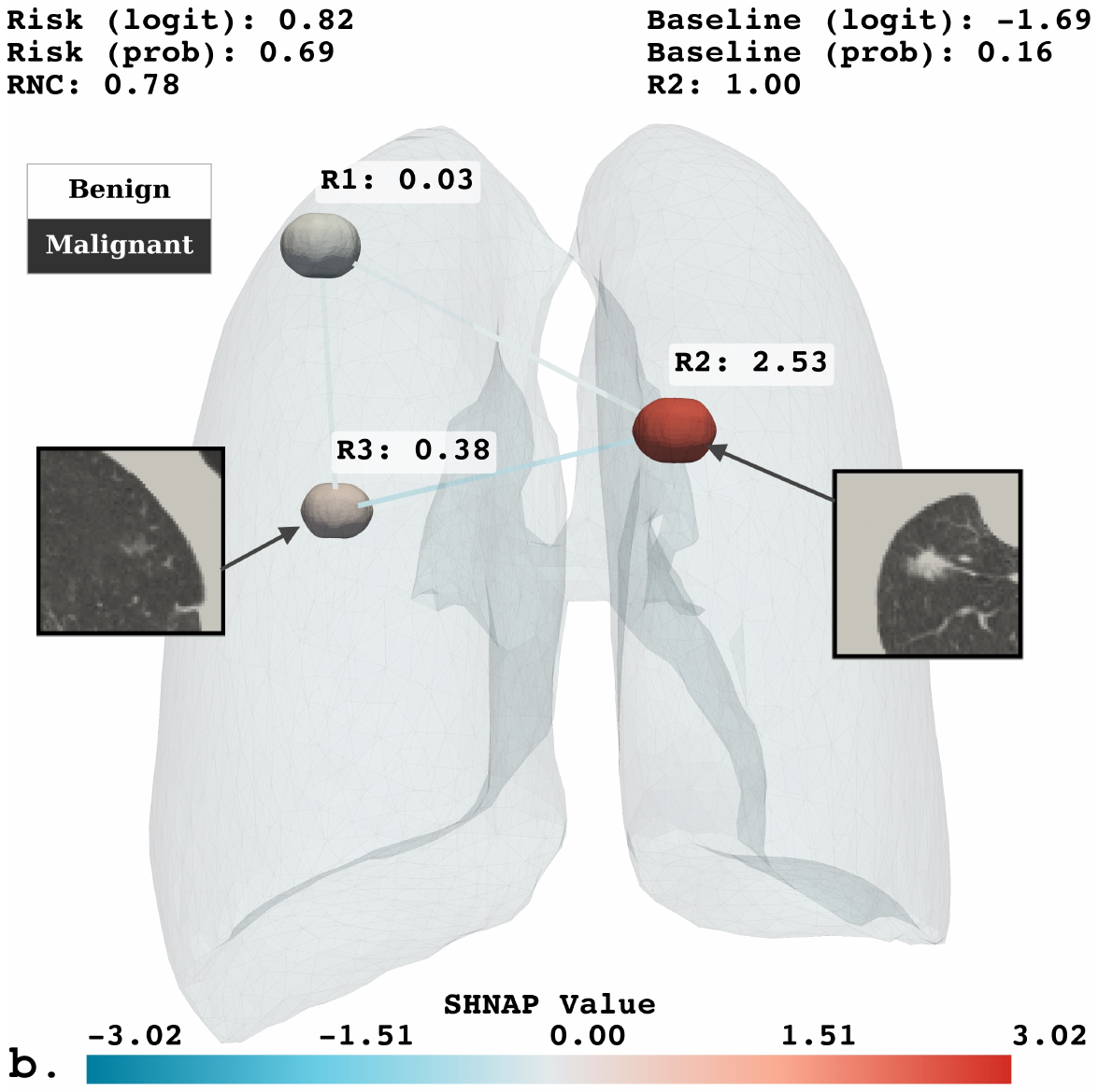}
    \includegraphics{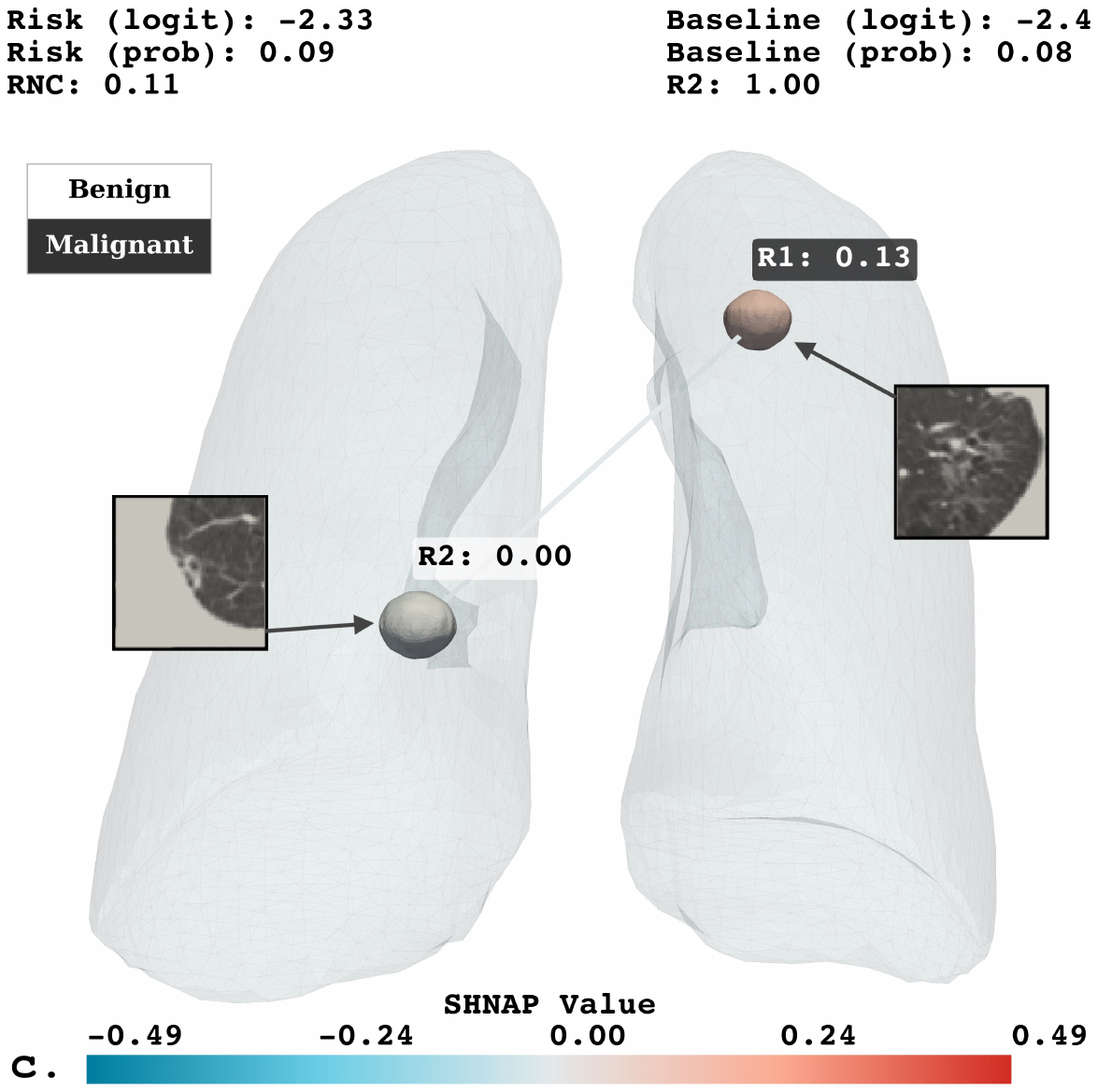}
    
    \vspace{-6em} 
    
    \includegraphics{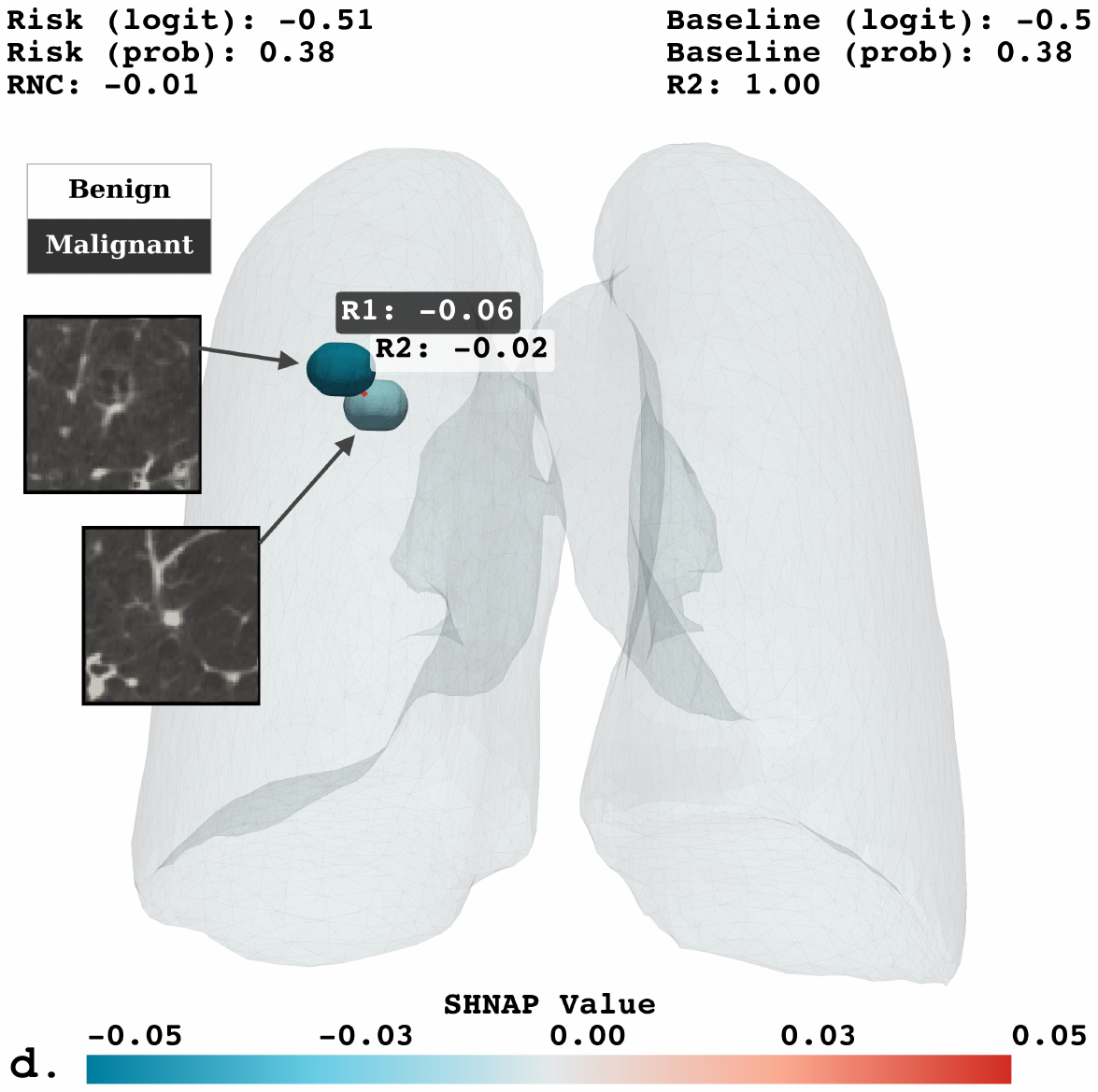}
    \includegraphics{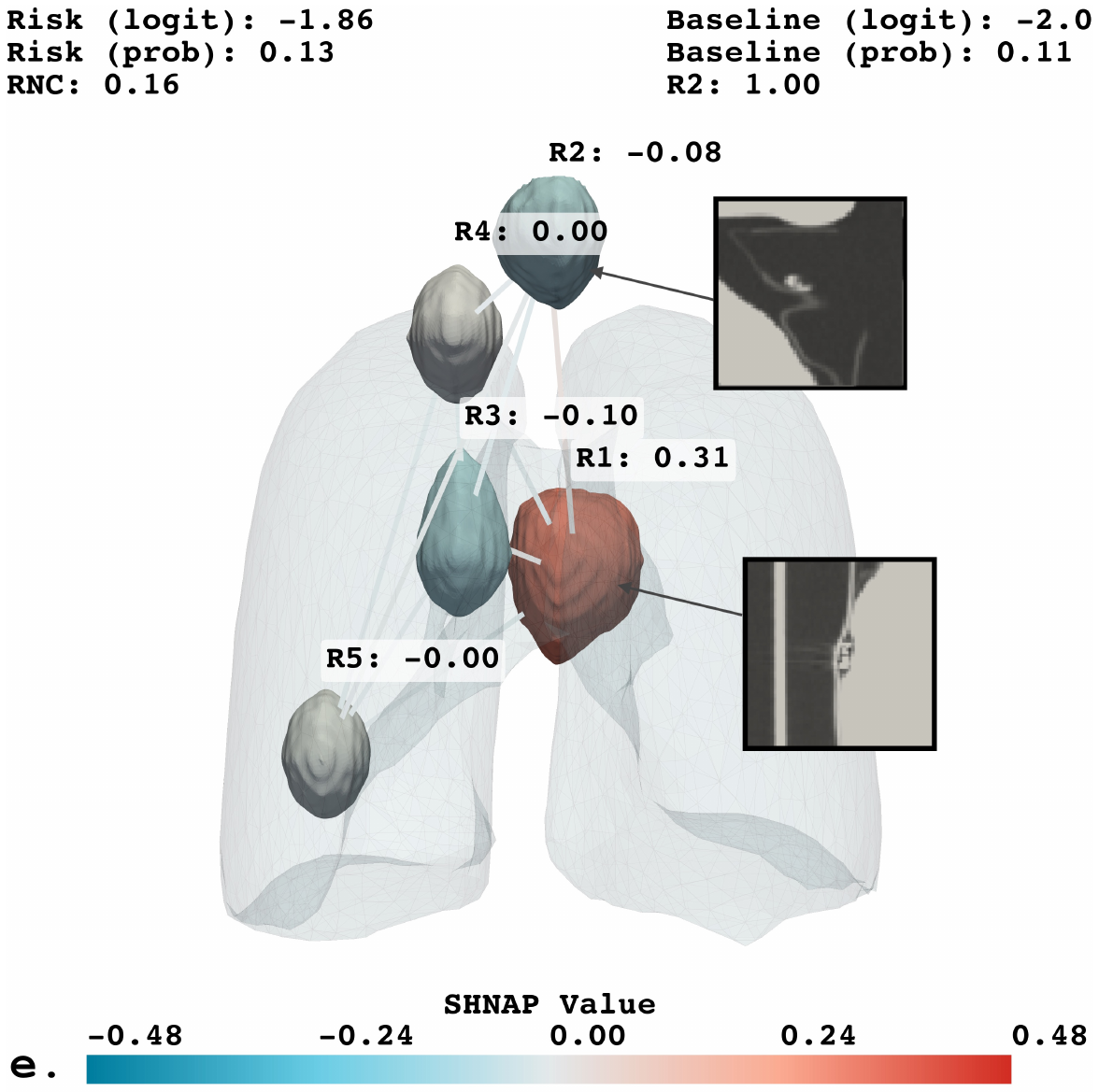}
    \includegraphics{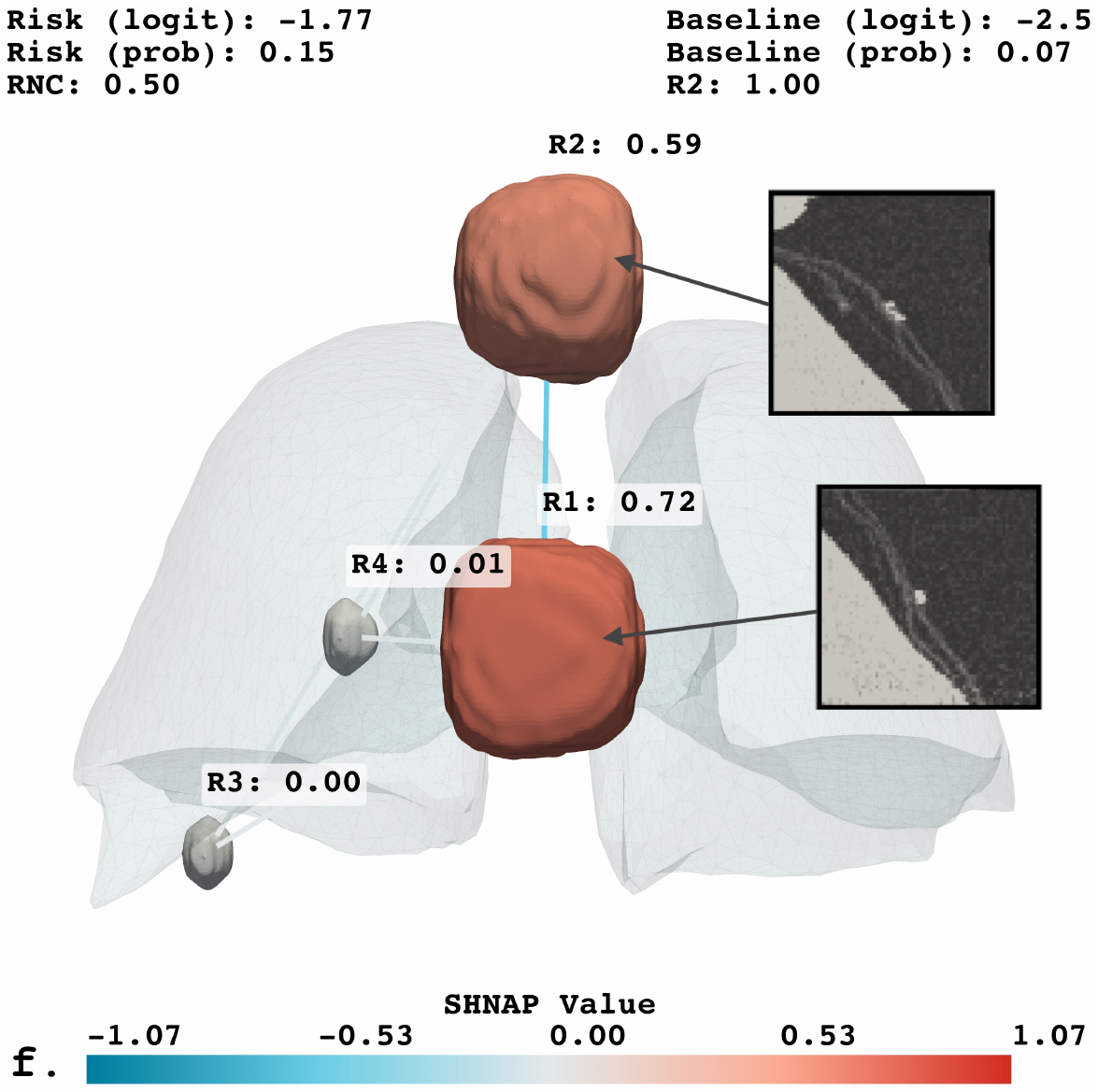}
    
    \vspace{-3em} 
    \caption{\nshap explanations for Sybil predictions across various patients. Each subplot indicates the initial prediction, the baseline term, local fidelity ($R^2$), and \her. 3D renders display attribution magnitudes, with labels for benign and malignant nodules.}
    \vspace{-1em}
    \label{fig:nshap_nodules}
\end{figure*}

\Cref{fig:nshap_statistics} shows \her densities for the LUNA25 test split (\textbf{a.}) and iLDCT (\textbf{b.}). On LUNA25, the distribution for benign cases is highly right-skewed. Intuitively, low nodule contribution aligns with clinical reasoning: absent pathology, risk estimates should rely on background markers like emphysema. The long positive tail thus signals potential flaws where benign nodules erroneously drive high risk. Conversely, confirmed cancer cases in this dataset exhibit a bimodal distribution with a significant mode near 0, revealing that Sybil frequently ignores known lesions in favor of background context. In iLDCT (\textbf{b.}), characterized by higher severity, Sybil's focus shifts even more toward nodules, highlighting its capacity to identify ominous lesions. Yet, this increased sensitivity coincides with a larger tail of erroneous high attributions for benign findings.

\textbf{Understanding false positives.} \Cref{fig:nshap_nodules} (\textbf{a.}) visualizes Sybil's reasoning on a false positive where nodules alone drive $60\%$ of the risk. Among 5 nodules, Sybil focuses almost entirely on a large, dense lesion in the upper right lung (\textbf{R3}), ignoring two smaller ones in the left lung (\textbf{R1}, \textbf{R2}). This behavior mimics a radiologist's caution: \textbf{R3} is highly suspicious, while the global spread of smaller nodules suggests prior infection rather than malignancy. Sybil's caution overshoots the ground truth (which benefits from multi-year stability data unavailable to the model), but the decision logic is understandable.

\Cref{fig:nshap_nodules} (\textbf{b.}) reveals another false positive where $78\%$ of the risk stems from nodules, dominated by a single lesion (\textbf{R2}) with a dense center and ground-glass opacity. While this morphology can indicate invasive adenocarcinoma, it also typifies pneumonia. Sybil's high risk estimate reflects justifiable caution. However, inspection of three-year longitudinal follow-up (\cref{fig:shnap_extended_b_1,fig:shnap_extended_b_2}) reveals dangerous instability: the model's focus shifts to a different nodule (\textbf{R1}), neglecting \textbf{R2}. While \textbf{R1} is suspicious, the reasoning \emph{shift} exposes inconsistency, rendering the model untrustworthy despite the initial defensible prediction.

\textbf{False negatives reveal flawed reasoning.} \Cref{fig:nshap_nodules} (\textbf{c.}) presents a malignant case where Sybil severely underestimates risk. Although it correctly ranks the spiculated nodule (\textbf{R1}) above the pleural one (\textbf{R2}), the total nodule contribution is merely $11\%$. Crucially, both are subpleural, a common site for adenocarcinoma, yet the model fails to assign them significant weight, indicating a potential suppression mechanism. \Cref{fig:shnap_extended_c_1,fig:shnap_extended_c_2} further illustrates this systematic failure, where Sybil repeatedly ignores a malignant pleural nodule across consecutive yearly scans.

\textbf{Right for wrong reasons.} \Cref{fig:nshap_nodules} (\textbf{d.}) demonstrates that correct predictions do not imply correct reasoning. In this malignant case, Sybil assigns negative attribution to the actual nodules, effectively treating them as evidence \emph{against} cancer. It is only their interaction that negates this effect, rendering their total contribution as effectively zero. Consequently, the correct high-risk prediction is driven almost entirely by background features, while the primary evidence, distinct malignant patterns, is ignored.

\textbf{Beyond pulmonary nodules.} \nshap provides rigorous insights into reasoning about nodules, but model developers must also uncover reliance on out-of-distribution patterns like artifacts and spurious correlations \citep{pmlrv235}. To shift perspective, we propose \emph{generalized} \nshap (\gnshap), which views Sybil as an \lmpi over arbitrary image regions by replacing nodule indicators with region indicators $z_i$. This leverages SDB to replace specific areas with their ``most likely'' synthesized counterparts, effectively scrubbing rare artifacts. A key challenge is choosing these regions. While prior work on counterfactuals employs post-hoc attributions to detect important areas \citep{sobieski2025rethinking}, we leverage Sybil's own attention mechanism. By binarizing attention maps, we directly probe the role of salient sub-volumes in the decision process.

\textbf{Discovering clinically unjustified artifacts.} We explore LUNA25 and iLDCT using the relative contribution of attention-based regions (see \cref{app:nshap_statistics_attention}). Our primary findings reveal that attention often points to regions with non-zero \gnshap attributions that do not overlap with pulmonary nodules. \Cref{fig:nshap_nodules} (\textbf{e.}) displays 5 distinct attention-based regions, 2 of which are located \emph{outside} the patient's body. Dominant \textbf{R1} points to an artifact resembling metal snaps used to close a hospital gown, pressed between the patient's skin and the scanner table. Moreover, while contributing negatively to the prediction, \textbf{R2} points to a cross-section of the patient's chin (mandible), suggesting that Sybil treats it as a large, solitary mass resembling a benign nodule.

\Cref{fig:nshap_nodules} (\textbf{f.}) presents a critical failure in a benign case where $50\%$ of the predicted risk stems from the joint influence of two symmetric objects outside the patient's body (\textbf{R1} and \textbf{R2}). Upon closer inspection, these regions reveal ECG electrodes attached to the chest skin. While likely used for cardiac synchronization to avoid motion blur, Sybil appears to correlate these leads with critical conditions, erroneously inflating the risk. Such spurious correlations, akin to ``hospital tag'' shortcuts in other domains \citep{BANIECKI2025103026,mishra2016overview}, make deployment unacceptable. Further examples in \cref{app:shnap_extended} show reliance on other clinically unjustified artifacts, such as thyroid goiters or metallic objects outside the body.

\textbf{Influential regions are sparse.} The richness of findings based on \nshap and \gnshap raises the question of whether Sybil is sensitive to perturbations in \emph{any} arbitrary region. To ablate this, we generate \gnshap explanations for random regions within the lung volume that are disjoint from both nodules and attention-based regions. \Cref{fig:nshap_backgrounds} shows that their importance is highly concentrated around zero, confirming that influential regions are sparse and specific, rather than the model simply reacting to random changes.
\vspace{-1em}
\subsection{Opening Sybil's black box with \ninject}
\begin{figure}[htbp]
    \centering
    \includegraphics[width=0.98\linewidth]{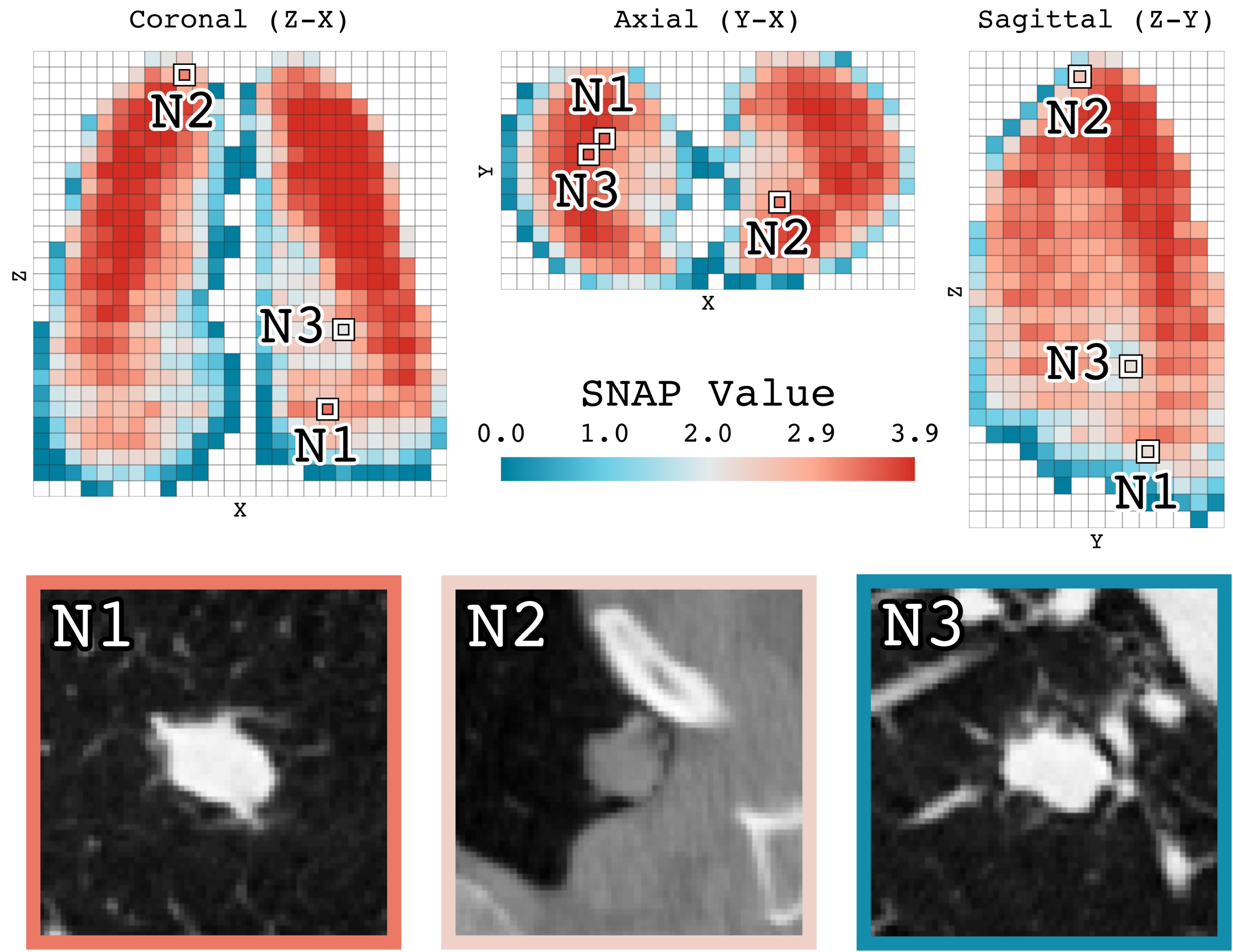}
    \caption{\ninject attribution map across three anatomical views, featuring example nodule insertions. Each nodule's value range is optimized for visual clarity.}
    \vspace{-1.5em}
    \label{fig:ninject_heatmaps_high}
\end{figure}
\textbf{Revealing local failures.} \Cref{fig:ninject_heatmaps_high} visualizes a high-resolution \ninject map of over 5,000 insertions of a malignant nodule in a single patient. Three specific sites highlight Sybil's variable sensitivity: \textbf{N1} (lung base) triggers a strong response, indicating correct identification; \textbf{N2} (near the pleura) shows a weaker response, suggesting distraction; and \textbf{N3} reveals a complete failure to detect the nodule, likely lost by Sybil in surrounding tissue. These examples demonstrate that while Sybil's risk estimates are generally smooth, sensitivity is not uniform and can fail locally.

\textbf{Revealing global anatomical biases.} To explore this at scale, we generated 240 patient-nodule combinations (20 nodules $\times$ 12 scans), performing $\approx 900$ insertions per combination (total $\approx 200,000$ samples), see \cref{app:ninject_attribution_maps}. We aggregated attributions by lung lobe and performed a two-way ANOVA. Results show significant main effects for patient identity ($p < 0.001$) and lobe class ($p < 0.001$), confirming that some patients naturally trigger higher risks and that anatomical sensitivity varies distinctively. Crucially, the interaction between patient and lobe was insignificant ($p \approx 1.0$), indicating that lobar bias is a \emph{global characteristic} of Sybil, independent of patient-specific variation.
\begin{figure}[b]
\vspace{-1em}
    \centering
    \includegraphics[width=0.99\linewidth]{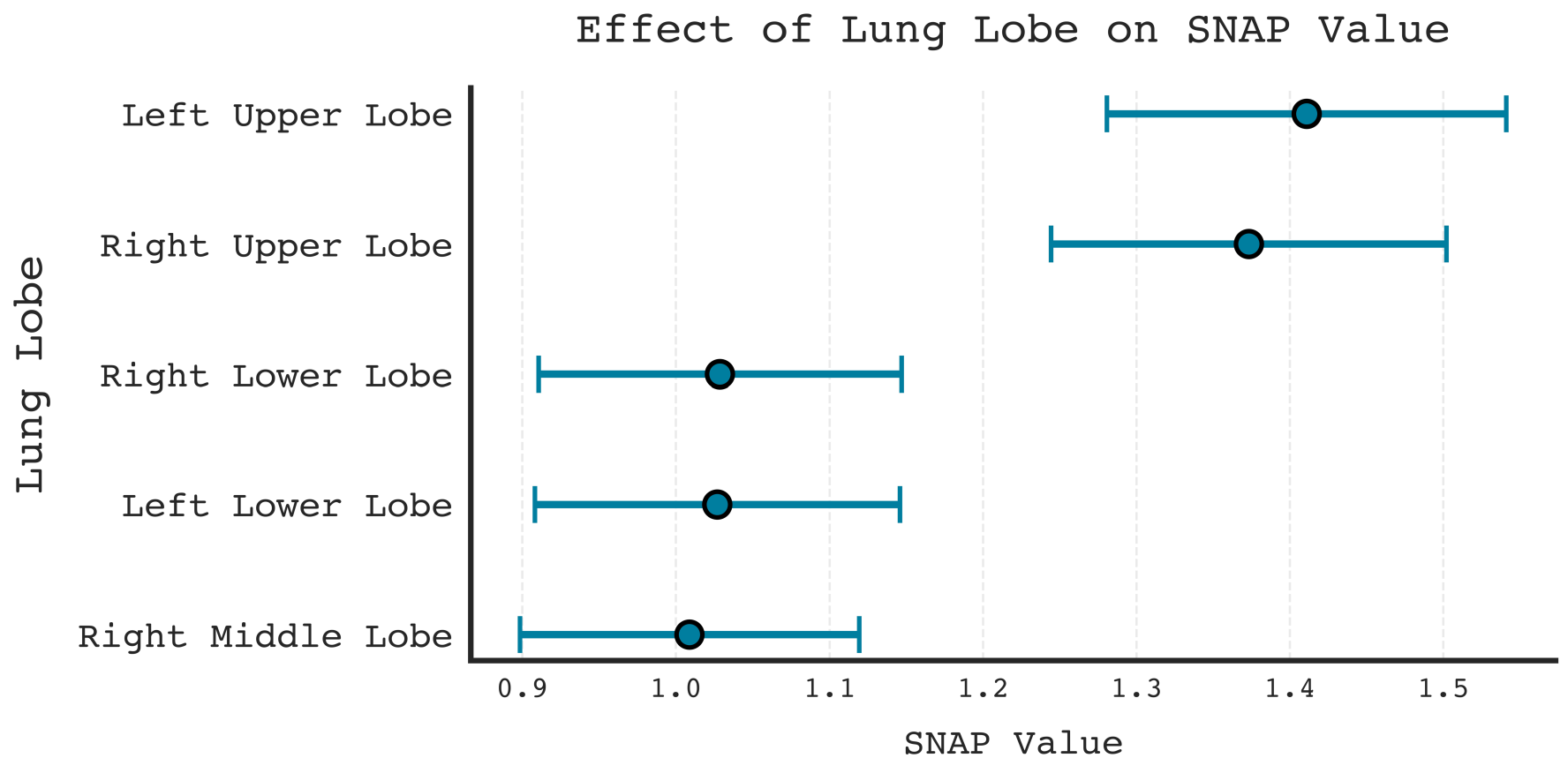}
    \caption{Average \ninject across 240 patient–nodule combinations, stratified by anatomical lung lobe.}
    \vspace{-1em}
    \label{fig:tukey}
\end{figure}
To determine directionality, we performed a post-hoc Tukey’s HSD test (\citealp{tukey1949comparing}, \cref{fig:tukey}). Attribution in the left and right upper lobes was significantly higher than in the middle/lower lobes ($p \le 0.009$). This aligns perfectly with clinical gold standards (\eg, PanCan, Mayo models \citep{swensen1997probability,mcwilliams2013probability}), which identify upper lobe location as a significant predictor of malignancy. Furthermore, the right middle lobe was indistinguishable from the lower lobes ($p-\text{value}\approx1.0$), forming a single inferior low-attribution zone, also consistent with literature. Finally, Sybil correctly ignores laterality (left vs.\ right), treating them as equivalent.

\textbf{Radial sensitivity bias.} While lobar biases align with medical knowledge, \ninject also reveals misalignment. Building on our \nshap findings of false negatives near the pleura (\cref{fig:nshap_nodules}), we identified a systematic failure to detect peripheral nodules. A linear regression predicting attribution from the distance-to-pleura yielded a significant positive coefficient ($p < 0.001$), indicating that sensitivity drops near the lung boundary. While distance alone explained little variance ($R^2=0.071$), adding interactions with nodule identity raised this to $R^2=0.455$, confirming a heterogeneous effect where malignant nodules are progressively attenuated near the boundary while benign ones remain robust. For an example trend, see \cref{fig:snap_radial_bias}.

We hypothesize this \emph{radial sensitivity bias} stems from zero-padding in 3D convolutions, a known cause of activation attenuation at boundaries \citep{alsallakh2020mind}. This structural vulnerability is clinically concerning: adenocarcinoma, the most common lung cancer subtype, predominantly arises in the periphery \citep{travis2011international}, making this blind spot a critical failure mode.
\vspace{-1em}
\subsection{Analyzing the background effect}
The decomposition in \cref{eq:shnap} enables granular analysis by isolating the background effect (the nodule-removed scan). While not our primary focus, analyzing this baseline offers complementary insights. Preliminary mixed-effects regression links the baseline term $\mu_\mathbf{x}$ to patient age, showing a positive trend on LUNA25 ($p=0.05$) and iLDCT ($p=0.027$). This implies Sybil likely infers age from global cues like bone density, embedding it into the baseline risk. These findings suggest that global features hold strong predictive power, justifying future work to disentangle them from localized pathologies.
\vspace{-1em}
\section{Discussion and limitations}
\label{sec:discussion_limitations}
We audited Sybil, a deep-learning lung cancer risk model, through the \nall framework. While checking for correct predictions is standard, our principled analysis of decision mechanisms corroborated Sybil's discriminative power and exposed critical reasoning misalignments. Our approach answers recent calls to prioritize strict model verification over purely observational studies \citep{biecek2025model}.

A primary limitation is the reliance on partially synthetic data, introducing the risk of generative artifacts. We mitigated this via a blinded expert study, although the pursuit of provably robust counterfactuals remains an active frontier \citep{anonymous2025counterfactual}. Crucially, \nall is model-agnostic, relying exclusively on input-output pairs. This allows the framework to be applied to arbitrary systems, including proprietary commercial models like Optellum \citep{Massion2022Optellum}, highlighting the immense value of domain-specific, object-level explanations.

\newpage
\section*{Impact Statement}

This paper advances the field of model explainability within the high-stakes domain of lung cancer risk prediction. While machine learning models demonstrate significant potential for automating lung cancer screening, our work highlights the necessity of rigorous auditing prior to clinical deployment. We demonstrate that reliance on clinically unjustified artifacts or spatial biases can lead to unreliable predictions. By providing a framework for generative interventions, this work contributes to the ethical development of AI in medicine, promoting transparency and safety. The societal implications include fostering trust in automated systems that patients and professionals rely on, ensuring that diagnostic assessments are both interpretable and accountable.

\section*{Acknowledgments}

Work on this project is financially supported by the Polish National Science Centre PRELUDIUM BIS grant No. \texttt{2023/50/O/ST6/00301} and the Foundation for Polish Science (FNP) grant ‘Centre for Credible AI’ No. \texttt{FENG.02.01-IP.05-0058/24}.

The computational resources for this work were provided by the Laboratory of Bioinformatics and Computational Genomics and the High Performance Computing Center of the Faculty of Mathematics and Information Science, Warsaw University of Technology. We also gratefully acknowledge Poland's High-performance Infrastructure PLGrid ACC Cyfronet AGH for providing computer facilities and support within computational grant no. \texttt{PLG/2025/018330}.

Finally, we express our gratitude to Hanna Piotrowska for her valuable assistance in preparing the visualizations and to Zuzanna Matuszewska for the detailed annotations of initial results.

\bibliography{main}
\bibliographystyle{icml2026}

\newpage
\appendix
\onecolumn

\startcontents[sections]
\printcontents[sections]{l}{1}{\setcounter{tocdepth}{2}}

\section{Extended related works}

\label{app:shap_nsv}
\subsection{SHAP and nSV.} While various methods exist to estimate the coefficients of an additive model, SHAP \citep{lundberg_shap} provides a unique solution that satisfies three distinct axiomatic properties. \textbf{P1.} \emph{Local accuracy} guarantees that the sum of the feature attributions matches the original model output when all features are present. \textbf{P2.} \emph{Missingness} requires that if a feature is absent from the input, its assigned attribution must be zero. \textbf{P3.} \emph{Consistency} ensures that if a model changes such that the marginal contribution of a feature increases or remains the same across all possible contexts, its attribution should not decrease. These properties ensure that the resulting explanations are faithful and allow for reliable comparison between different models.

Building on this additive framework, we extend univariate attribution to higher-order interactions through nSV \citep{bordt2023shapley}. While traditional SHAP decomposes the model output into individual feature contributions, nSV captures interaction effects between groups of features explicitly. This effectively represents the model as a Generalized Additive Model with interactions, where the output is the sum of main effects and their combinations. This decomposition is uniquely defined by a set of generalized axiomatic properties. \textbf{P1.} \emph{Efficiency} ensures that the total sum of all interaction effects, including a constant baseline value, perfectly recovers the full model output. \textbf{P2.} \emph{Symmetry} dictates that if two features contribute identically to every possible combination of other features, their respective interaction attributions must be equal. \textbf{P3.} \emph{Null interaction} requires that if a feature adds no value to any possible grouping, all interaction terms involving that feature are assigned an attribution of zero. Together, these axioms provide a theoretically grounded method for auditing complex models, ensuring that interaction effects are neither arbitrarily assigned nor omit critical dependencies.

\subsection{Generalized Additive Models.} Linear models stand at the core of statistics, offering a plethora of practical tools for explaining underlying data mechanisms, though often at the cost of predictive capacity \citep{hastie2009elements}. Their simplicity makes them widely recognized as inherently \emph{interpretable} \citep{rudin2019stop}. Generalized Additive Models (GAMs, \citealp{hastie1990generalized}) combine their white-box nature with the flexibility of non-linear models. Formally, a GAM of order $n$ is defined as a function $f: \mathbb{R}^d\rightarrow\mathbb{R}$ that can be written in the form
\begin{equation}\label{eq:gam}
    f(\mathbf{x}) = \sum_{S\subseteq [d], 0 \leq |S| \leq n} f_S(\mathbf{x}_S),
\end{equation}
where $[d]=\{1,\dots,d\}$ and $S=\{ s_1,\dots, s_k\}\subseteq[d], |S|=k$ is used to index both the data features $\mathbf{x}_S=(x_{s_1},\dots,x_{s_k})$ and the component functions acting solely on their subsets $f_S(\mathbf{x}_S)=f_{s_1,\dots, s_k} (x_{s_1},\dots,x_{s_k})$.


\subsection{Counterfactual explanations.} Standing at the top of Pearl's causality ladder \citep{Pearl_2009}, visual counterfactual explanations (VCEs) aim to modify a given sample in a minimal and semantically meaningful way that also changes the prediction of a decision model to a target one, allowing to explore its behavior under \emph{what-if} scenarios. To guide the image modification along the data manifold, most approaches make use of a separate generative model as an approximation to the data distribution. Commonly, finding VCEs is formulated as solving the optimization problem
\begin{equation}\label{eq:vce_objective}
    \arg \min_{\hat{\mathbf{x}}}{s(\mathbf{x}, \hat{\mathbf{x}}) - \lambda \cdot p_{\boldsymbol{\theta}} (y'\mid \hat{\mathbf{x}})},
\end{equation}
where $s(\mathbf{x}, \hat{\mathbf{x}})$ is the \emph{semantic} distance between the true $\mathbf{x}$ and modified $\hat{\mathbf{x}}$, $p_{\boldsymbol{\theta}}$ is the $\boldsymbol{\theta}$-parameterized decision model, $y'$ is the target decision and $\lambda>0$ controls the trade-off between the decision change and semantic distance.

Approaches for solving \cref{eq:vce_objective} may be categorized as either white- or black-box. The former differentiate through \cref{eq:vce_objective} directly, utilizing access to weights and computational graph of the decision model \citep{jacob2022steex,jeanneret2022diffusion,augustin2022diffusion,jeanneret2023adversarial,augustin2024digin,sobieski2025rethinking}. The latter assume more general scenario of access limited to only input-output interaction, making the optimization more difficult. This can be performed either explicitly by approximating \cref{eq:vce_objective} \citep{sobieski2024global,jeanneret2024text} or implicitly by observing its behavior under modification of various semantic attributes \citep{kazimi2024explaining}.

\subsection{XAI in medical imaging.}\label{app:xai_medical}
Prior approaches for understanding the decision-making mechanisms of predictive models trained on medical images (e.g., CT, MRI, X-ray) were largely limited to the direct application of standard XAI techniques \citep{ahmed2026explainable}. In the context of Grad-CAM \citep{selvaraju2020grad}, \citet{PANWAR2020110190}, \citet{Moujahid2022}, \citet{Musthafa2024}, and \citet{Hammad2025} apply it to localize pathologies in lung CT, chest X-rays, and brain MRI, while \citet{Panboonyuen2026} critically analyzes its reliability. \citet{Nahiduzzaman2024} and \citet{Rai2025} use SHAP to construct lung cancer detection frameworks. \citet{Eitel2019} apply Layer-Wise Relevance Propagation (LRP, \citealp{bach2015pixel}) to MRI-based multiple sclerosis diagnosis. \citet{Sigut2023} approximate CNN predictions using surrogate models for glaucoma diagnosis from fundus images. \citet{Elbouknify2023}, \citet{Islam2025}, and \citet{Navaratnarajah2025} combine multiple XAI methods to obtain a more comprehensive view of the diagnosis across different modalities. In a more sophisticated fashion, \citet{mertes2022ganterfactual}, \citet{Arora2024}, and \citet{Navaratnarajah2025} analyze counterfactual scenarios in mammography and neuroimaging contexts.

Crucially, while prior approaches reuse or adapt standard XAI techniques, they do not consider explanations tailored specifically to the 3D medical imaging domain, which presents its own set of challenges and characteristic predictive features, a gap that is the focus of our work. Moreover, our work goes beyond standard approaches by leveraging recent advances in generative modeling to reinterpret the decision-making process within a simplified feature space composed of pulmonary nodules, upon which we construct explanations using state-of-the-art methods for Shapley-based Interaction Indices \citep{grabisch1999axiomatic}.

\subsection{Synthesis, nodule removal and insertion.}\label{app:removal_insertion}
3D medical imaging poses significant computational challenges for synthesis due to large image sizes, complex anatomical structures, and the importance of high-frequency details. The focus of recent approaches has shifted almost entirely to diffusion models (DMs) due to their natural advantages in image generation quality \citep{dhariwal2021beat}. MedicalDiffusion \citep{khader2023denoising} was one of the first approaches to successfully apply the DM framework to 3D medical image synthesis. Subsequent models, such as Lung-DDPM \citep{Jiang2025LungDDPMSL}, Lung-DDPM+ \citep{JIANG2025111290}, and Med-DDPM \citep{10493074}, extended the standard framework with semantic-layout guidance, allowing for controlled generation with faster inference times. For the same purpose, GEM-3D \citep{zhu2024generative} utilized conditional diffusion models \citep{batzolis2021conditional}. LAND \citep{oliveras2025land} applied latent DMs \citep{rombach2022high} to generate high-quality 3D chest CT scans from anatomical masks. In the domain of image forensics, M3DSYNTH \citep{zingarini2024m3dsynth} introduced a DM capable of shrinking and enlarging existing pulmonary nodules. \citet{kim2025tumor} combined Generative Adversarial Networks and Brownian Bridge DMs to allow for the removal, synthesis, and modification of nodule shape and texture based on radiomics features. MAISI \citep{guo2025maisi} combined latent DMs with a volume compression network to generate high-resolution CT images using various conditioning mechanisms, while MAISIv2 \citep{zhao2025maisiv2accelerated3dhighresolution} extended this framework to rectified flows \citep{liuflow}.

In contrast to our work, prior approaches either focused on the pure synthesis of entire volumes without granular control over pulmonary nodules \citep{Jiang2025LungDDPMSL,JIANG2025111290,10493074,zhu2024generative} or required annotated data during training to enable nodule removal and synthesis \citep{oliveras2025land,zingarini2024m3dsynth,kim2025tumor,guo2025maisi,zhao2025maisiv2accelerated3dhighresolution}. Crucially, none of these works allows for both nodule removal \emph{and} the insertion of an existing, extracted nodule that remains aligned with the new context and preserves its original properties, all while relying solely on unannotated CT scans for training.

\subsection{Diffusion models.} In recent years, generative modeling of visual data has been largely dominated by diffusion models \citep{ho2020denoising, dhariwal2021beat}, with a particular focus on their continuous-time formulation \citep{song2019generative,songscore}. Inspired by the physics of non-equilibrium thermodynamics \citep{sohl2015deep}, these models learn to reverse a noising process through iterative denoising. Formally, let $p(\mathbf{x}_0)$ represent the data and $p(\mathbf{x}_1)$ the prior distribution respectively. Diffusion models connect the two by considering $p(\mathbf{x}_t)$ for $t\in[0, 1]$ as the distribution induced by the \emph{forward process} evolving according to a stochastic differential equation (SDE)
\begin{equation}
    \diff \mathbf{x}_t = \ldriftcoef \diff t + \ldiffcoef \diff \wiener,\label{eq:forward_diffusion_}
\end{equation}
with a \emph{linear} drift term $\ldriftcoef$, a time-dependent matrix $\mathbf{F}_t \in \mathbb{R}^{d\times d}$ and matrix-valued diffusion coefficient $\ldiffcoef$. \Cref{eq:forward_diffusion} allows for mapping the data distribution to an easy-to-sample prior, \eg, an isotropic zero-mean Gaussian. To invert this mapping, a \emph{reverse process} is considered
\begin{equation}
    \diff \mathbf{x}_t = [\ldriftcoef - \ldiffcoef \ldiffcoef^\top \score] \diff t + \ldiffcoef \diff \rwiener, \label{eq:reverse_diffusion_}
\end{equation}
where $\score$ is the \emph{score function}. The inversion is understood as the marginals $p(\mathbf{x}_t)$ of \cref{eq:forward_diffusion} and \cref{eq:reverse_diffusion} being equal. As $\score$ is unavailable in most practical scenarios, a neural network $\mathbf{s}_{\boldsymbol{\xi}}$ is trained to regress $\scorec$ marginalized over samples $\mathbf{x}_0$, leading to an asymptotically unbiased estimate of $\score$ \citep{JMLR:v6:hyvarinen05a}. The choice of the drift and diffusion coefficients $\mathbf{F}_t, \mathbf{G}_t$ determines the dynamics of the mapping between the Gaussian and data distribution. A large majority of works restricted them to simple scalar functions, \ie, $\mathbf{F}_t=f_t \mathbf{I}, \mathbf{G}_t=g_t \mathbf{I}$ for some time-dependent $f_t, g_t \in \mathcal{C} ([0,1], \mathbb{R})$ \citep{ho2020denoising,songscore,dhariwal2021beat,song2021denoising}.

\subsection{Diffusion bridges.} An important extension of diffusion models are the so-called \emph{diffusion bridges}, which generalize to mappings between arbitrary distributions for $p(\mathbf{x}_0)$ and $p(\mathbf{x}_1)$. Initial works arrived at such mappings by considering more general SDEs with additional dependence on the ``endpoints'' from $p(\mathbf{x}_1)$ \citep{luo2023image,liu20232,yue2024image,zhoudenoising,li2023bbdm,liu20232,luo2023image,yue2024image,zhoudenoising,kim2024unpaired,de2024schrodinger}. Diffusion bridges offer a principled data-based approach to solving \emph{inverse problems}, \ie, tasks of the form 
\begin{equation}\label{eq:general_inverse_problem}
\mathbf{y}=\mathcal{A} (\mathbf{x})    
\end{equation}
with $\mathbf{x}$ being the true signal, $\mathcal{A}$ a (possibly stochastic) measurement operator, $\mathbf{y}$ the resulting measurement, with the goal of retrieving the true $\mathbf{x}$ from the observed $\mathbf{y}$. Attempting to solve an inverse problem with a diffusion bridge can be generally described as setting $p(\mathbf{x}_0)=p(\mathbf{x}), p(\mathbf{x}_1)=p(\mathbf{y})$ and learning its corresponding SDE using pairs $(\mathbf{x}, \mathbf{y})$ sampled from the joint $p(\mathbf{x}, \mathbf{y})$ induced by \cref{eq:general_inverse_problem}.
An important special case of \cref{eq:general_inverse_problem}, covering many practical applications, is the \emph{linear Gaussian} form
\begin{equation}\label{eq:linear_gaussian_inverse_problem}
    \mathbf{y}=\mathbf{A}\mathbf{x} + \boldsymbol{\Sigma}^{\frac{1}{2}}\boldsymbol{\varepsilon},
\end{equation}
where $\mathbf{A}\in\mathbb{R}^{m \times n}$ for $m,n\in\mathbb{N}$, $\boldsymbol{\varepsilon}\sim\mathcal{N} (\mathbf{0}_{m}, \mathbf{I}_{m\times m})$ and $\boldsymbol{\Sigma}\in\mathbb{R}_{\geq 0}^{m\times m}$ is a positive semi-definite covariance matrix. Notably, \citep{sobieski2025sdb} showed that by utilizing the matrix-valued formulation of \cref{eq:forward_diffusion,eq:reverse_diffusion}, where the drift and diffusion coefficients do not reduce to scalar functions, one may embed the system from \cref{eq:linear_gaussian_inverse_problem} directly into the SDE, leading to a Markovian diffusion bridge between $p(\mathbf{x})$ and $p(\mathbf{y})$ termed SDB.

For the purpose of this work, consider the specific form of \cref{eq:linear_gaussian_inverse_problem} for the no-noise inpainting problem with $\mathbf{y}=\mathbf{A}\mathbf{x}$, where $\mathbf{A}$ is the masking operator. Let $\mathbf{F}_t=\frac{\diff}{\diff t}\log{\alpha_t} (\mathbf{I} - \mathbf{A}^+\mathbf{A}), \mathbf{G}_t\mathbf{G}_t^\top=\left( \frac{\diff\beta_t}{\diff t}-2\beta_t\frac{\diff}{\diff t}\log{\alpha_t}\right)(\mathbf{I} - \mathbf{A}^+\mathbf{A})$ with $\mathbf{A}^+$ denoting the pseudoinverse of $\mathbf{A}$ and scalar functions $\alpha_t, \beta_t$ such that $\alpha_0 = 1, \alpha_1=0$ and $\beta_t \rightarrow 0$ uniformly for all $t$. These lead to SDB (SB), the Schr{\"o}dinger Bridge (SB) variant of SDB, which solves the optimal transport plan \citep{mikami2004monge} in the null space of $\mathbf{M}$, following the main theoretical result from \cite{liu20232}, while entirely preserving the range space component.

\section{Extended methodology}

\subsection{Pseudocode}

We include the pseudocode for both \nshap and \ninject in \cref{alg:shnap} and \cref{alg:snap} respectively.

\begin{algorithm}[h]
   \caption{SHNAP: SHapley Nodule Attribution Profiles}
   \label{alg:shnap}
\begin{algorithmic}[1]
   \STATE {\bfseries Input:} Original scan $\mathbf{x}$, logit function $f_{\boldsymbol{\theta}}$, score model $s_{\boldsymbol{\xi}}$, nodule set $\mathbf{N}$, nodule masks $\{\mathbf{A}_i\}_{i=1}^{|\mathbf{N}|}$
   \STATE {\bfseries Output:} Baseline $\mu_{\mathbf{x}}$, main effects $\phi_i$, pairwise interactions $\phi_{ij}$
   \FOR{each $i \in \mathbf{N}$}
      \STATE $\mathbf{x}_{1,i} \leftarrow \text{ForwardDiffusion} (\mathbf{x}, t=1, \text{mask}=\mathbf{A}_i)$ \hfill \# Remove nodule $i$
      \STATE $\tilde{\mathbf{x}}_i \leftarrow \text{ReverseSampling} (\mathbf{x}_{1,i}, s_{\boldsymbol{\xi}}, t=1 \to 0)$ \hfill \# Inpaint healthy tissue for $i$
   \ENDFOR
   \STATE $\mathbf{x}_{\emptyset} \leftarrow \text{Recompose} (\mathbf{x}, \{\tilde{\mathbf{x}}_i\}_{i \in \mathbf{N}})$ \hfill \# Construct nodule-free baseline
   \STATE $\mathcal{D} \leftarrow (\emptyset, f_{\boldsymbol{\theta}} (y_0 \mid \mathbf{x}_{\emptyset}))$ \hfill \# Initialize dataset
   \FOR{each subset $S \subseteq \mathbf{N}$}
      \STATE $\mathbf{x}_S \leftarrow \text{Recompose} (\mathbf{x}_{\emptyset}, \{\mathbf{x}_j\}_{j \in S})$ \hfill \# Reinsert original nodules from $S$
      \STATE $\mathcal{D} \leftarrow \mathcal{D} \cup (S, f_{\boldsymbol{\theta}} (y_0 \mid \mathbf{x}_S))$ \hfill \# Form dataset for nSV
   \ENDFOR
   \STATE Solve for $\mu_{\mathbf{x}}, \phi_i, \phi_{ij}$ with nSV \hfill \# Least-squares projection
   \STATE {\bfseries return} $\mu_{\mathbf{x}}, \phi_i, \phi_{ij}$
\end{algorithmic}
\end{algorithm}

\begin{algorithm}[h]
   \caption{SNAP: Substitutive Nodule Attribution Probing}
   \label{alg:snap}
\begin{algorithmic}[1]
   \STATE {\bfseries Input:} Target scan $\mathbf{x}$, source nodule content $\mathbf{r}$, mask $\mathbf{A}$, target coordinate $\mathbf{c}$, score model $s_{\boldsymbol{\xi}}$, logit function $f_{\boldsymbol{\theta}}$
   \STATE {\bfseries Output:} Attribution score $\psi_{\mathbf{c}}$
   \STATE {\bfseries Parameter:} Blending timestep $\tau = 0.3$
   \STATE Translate $\mathbf{r}$ and $\mathbf{A}$ to coordinate $\mathbf{c}$ \hfill \# Spatial alignment
   \STATE $\mathbf{x}_{paste} \leftarrow \text{CopyPaste} (\mathbf{x}, \mathbf{r}, \mathbf{A}, \mathbf{c})$ \hfill \# Simulate distribution $p(\mathbf{x}^2)$
   \STATE $\mathbf{x}_{\tau} \leftarrow \text{ForwardDiffusion} (\mathbf{x}_{paste}, t=\tau)$ \hfill \# Diffuse up to step $\tau$
   \STATE $\mathbf{x}_{\mathbf{c} \leftarrow \mathbf{r}} \leftarrow \text{ReverseSampling} (\mathbf{x}_{\tau}, s_{\boldsymbol{\xi}}, t=\tau \to 0)$ \hfill \# Align with the background
   \STATE $\psi_{\mathbf{c}} = f_{\boldsymbol{\theta}} (y_0 \mid \mathbf{x}_{\mathbf{c} \leftarrow \mathbf{r}}) - f_{\boldsymbol{\theta}} (y_0 \mid \mathbf{x})$ \hfill \# Compute log-odds ratio
   \STATE {\bfseries return} $\psi_{\mathbf{c}}$
\end{algorithmic}
\end{algorithm}

\section{Extended experiments}

\subsection{Risk correlation}

\begin{figure}[t]
    \centering
    \includegraphics[width=0.95\linewidth]{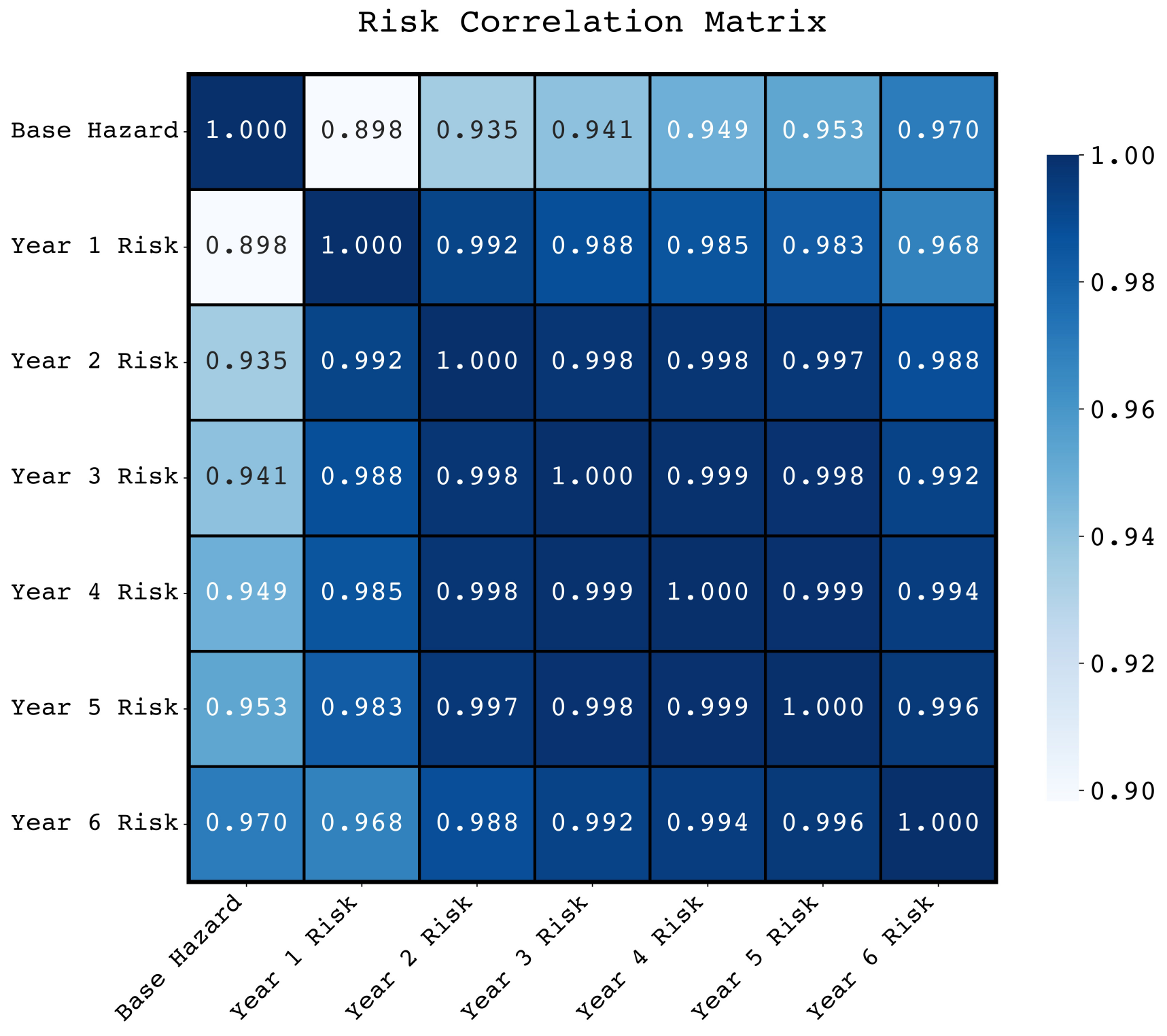}
    \caption{Correlation matrix of Sybil's base hazard and per-year risk outputs computed on iLDCT.}
    \label{fig:risk_correlation}
\end{figure}

\Cref{fig:risk_correlation} presents the correlation matrix between Sybil's base hazard and its per-year risk outputs as computed on iLDCT. The lowest observed correlation is $\approx 0.9$ between the first-year risk and the base hazard; this high degree of alignment suggests that the base risk serves as a robust and universal representation of the model's overall diagnostic output.

\subsection{Experimental Setup}
\label{app:experimental_setup}

\subsection{Datasets}
We base our analysis on three 3D Low-Dose Computed Tomography (LDCT) datasets, each serving a distinct role in auditing the model.

\textbf{D1. NLST.} We utilize the National Lung Screening Trial (NLST, \citealp{NLST2011}) using the official splits from Sybil. This large-scale cohort comprises over 28,000 training and 6,000 test scans. The data provides a high-variance backdrop for training, with each scan containing between 120 and 400 slices of $512^2$ pixel resolution and a slice thickness of $\leq 2.5$ mm.

\textbf{D2. LUNA25.} The LUng Nodule Analysis dataset (LUNA25, \citealp{Peeters2025LUNA}) is a curated subset of the NLST consisting of 4,069 scans from 2,120 patients. It is particularly valuable for our nodule-centric analysis as it includes precise coordinates for 555 malignant and 5,608 benign nodules. Crucially, the ground truth labels are highly reliable: malignancy was confirmed via biopsy (providing pathology data unavailable to visual inspection), while benignity was verified by stability over a 2-year follow-up period. We use the provided centroid coordinates to construct naive spherical segmentation masks for our interventions.

\textbf{D3. iLDCT.} To test robustness on out-of-distribution data, we employ an internal screening dataset (iLDCT) of 243 scans from 145 patients. Unlike the screening-focused NLST, this dataset features a higher prevalence of severe cases. It has been annotated by two expert radiologists who provided detailed segmentation masks and malignant/benign classifications, serving as a rigorous testbed for model reasoning under more pathological conditions.

\begin{figure}[h]
    \centering
    \includegraphics[width=0.95\linewidth]{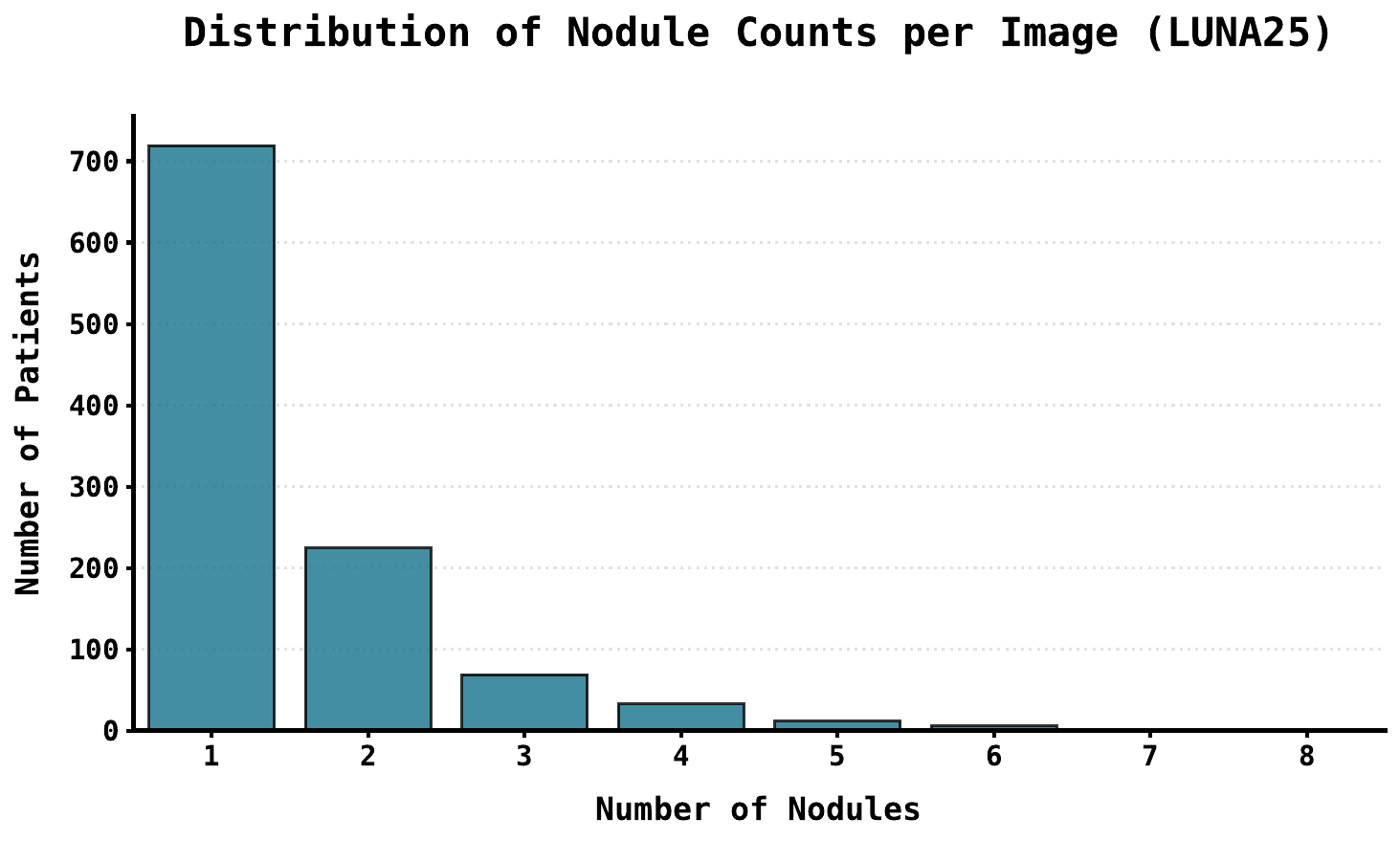}
    \caption{Distribution of nodule counts on LUNA25.}
    \label{fig:distribution_luna25}
\end{figure}

\begin{figure}[h]
    \centering
    \includegraphics[width=0.95\linewidth]{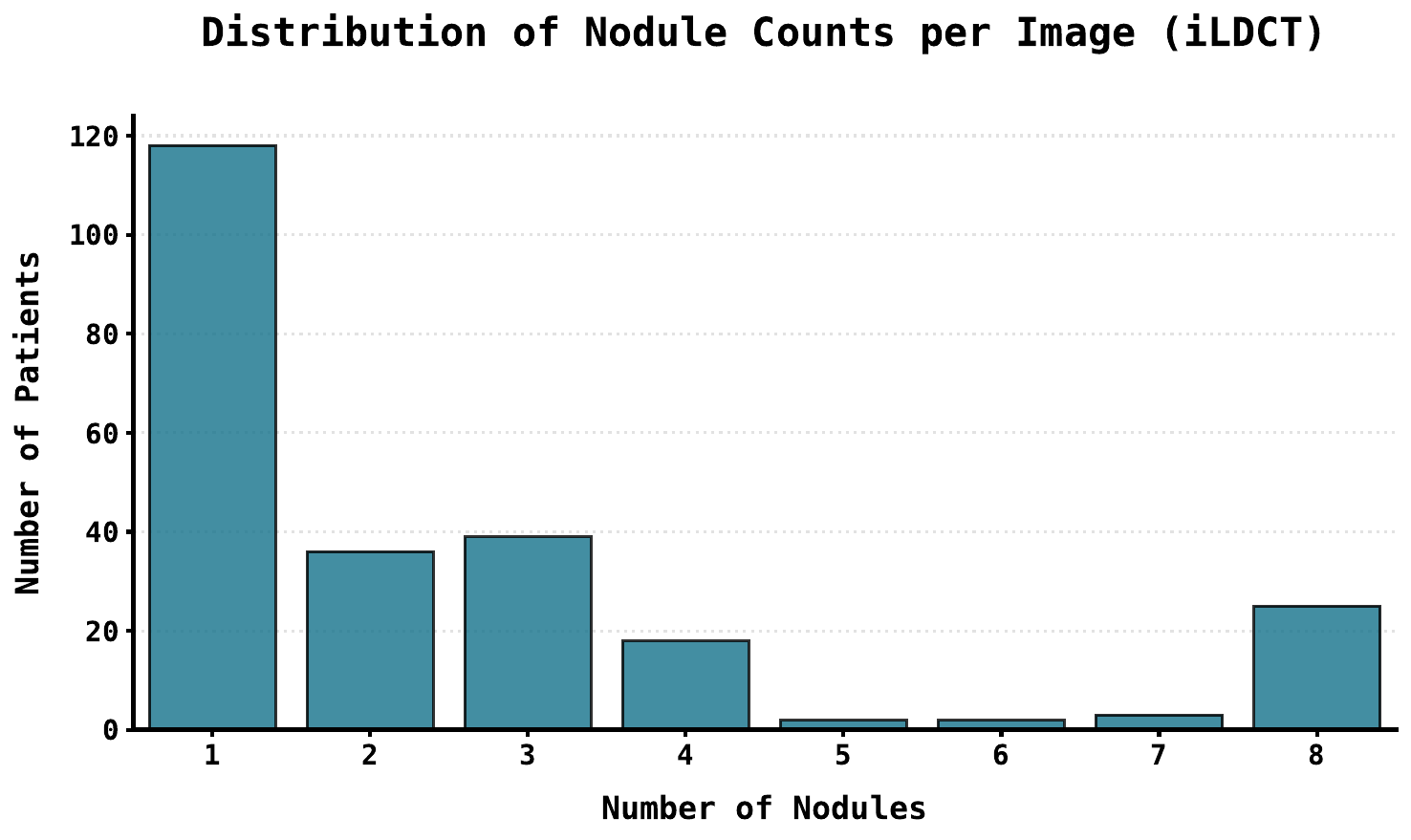}
    \caption{Distribution of nodule counts on iLDCT.}
    \label{fig:distribution_ildct}
\end{figure}

We include histograms for nodule counts for both LUNA25 and iLDCT in \cref{fig:distribution_luna25,fig:distribution_ildct}.

\subsection{SDB Training and Implementation}
\textbf{Computational Strategy.} Processing full 3D LDCT volumes is computationally prohibitive. We mitigate this by training SDB on randomly sampled $64^3$ cubes from volumes which were resampled to a target spacing of (1mm, 1mm, 2mm), utilizing the fact that nodule structure depends primarily on local high-frequency details rather than global context.

\textbf{Architecture and Training.} We employ a discrete-time (1000 steps) Schr{\"o}dinger Bridge (SB) variant of SDB. Our implementation adapts the code from \citet{liu20232} by replacing 2D convolutions with 3D counterparts and implementing FlashAttention \citep{dao2022flashattention} in BF16 to improve efficiency, while the remaining computations are performed in FP32. The resulting model (76M parameters) was trained for approximately $1,600$ A100-hours on the NLST training split.

The architecture operates on $64^3$ volumes with a base width of $32$ channels and multipliers of $\{1, 2, 3, 4\}$ across levels. Each resolution level contains $3$ residual blocks, with single-head self-attention ($32$ channels per head) applied at the $16^3$ and $8^3$ resolutions. Training utilized an effective batch size of $128$, a dropout rate of $0.1$, exponential moving average (EMA) of weights ($\rho = 0.99$), and a noise schedule with $\beta_{\max}=1.0$.

\textbf{Procedural Masking.} To avoid the bottleneck of costly manual labeling for training data, we generate subvolume masks procedurally using the metaballs algorithm \citep{blinn1982generalization,zhao2021large}. To further augment the variety of generated masks we randomly sample from 4 to 7 core-points, than for all points in the volume we compute the sum of cartesian distances to the core-points and binarize the mask using a randomly selected quintile (from near $0$ to $0.4$) as a threshold. Which results in random smooth regions with at most $40\%$ of the cube's volume covered.
This strategy yields a model capable of high-fidelity nodule removal and insertion without assuming training-time access to annotated nodule locations. This advantage enables large-scale training on the full NLST cohort, unlike recent methods that are limited to small, fully annotated cohorts \citep{zingarini2024m3dsynth,kim2025tumor}.

\textbf{Inference.} During the auditing phase, both nodule removal and insertion are performed using 100 Number of Function Evaluations (NFE) of the trained SDB model. To ensure that nodule insertions are precisely limited to the pulmonary volume, we utilize lung and lobe segmentation masks obtained from the TotalSegmentator \citep{wasserthal2023totalsegmentator} adaptation proposed by \citet{10943755}.

\subsection{Baseline comparison}

We compare SDB reconstruction capabilities against two pretrained models, Lung-DDPM \cite{Jiang2025LungDDPMSL} and MAISIv2 \cite{guo2025maisi}, adapted for inpainting via RePaint \cite{lugmayr2022repaint}, as well as pure noise and mean-value baselines. Quantitative results (see \Cref{tab:baseline_comparison}) were computed on 3,226 samples with metaball masks. To accommodate the pretrained baselines, inputs were padded to $128^3$ without mask resizing. Fréchet Inception Distance (FID) \citep{heusel2017gans} metrics were calculated using RadImageNet \cite{mei2022radimagenet} embeddings on slices containing masked regions.

\begin{table*}[h]
    \centering
    \small
    \caption{\textbf{Quantitative evaluation of inpainting methods.} 
    The table reports Fréchet Inception Distance (FID) across three planes (XY, XZ and YZ) and metrics computed on masked regions. 
    RMSE and MAE are reported in Hounsfield Units (HU).
    The proposed method is highlighted using a gray background.
    }
    \label{tbl:results_lungs}
    \begin{tblr}{
      colspec = {lccccccc},
      column{1} = {leftsep=10pt}, 
      column{Z} = {rightsep=10pt},
      vline{2,6} = {dashed},
      hline{2} = {solid},
      rowsep=3pt,
      row{7} = {bg=black!10}, 
    }
    \toprule
    & \SetCell[c=4]{c} \textbf{FID Metrics} ($\downarrow$) & & & & \SetCell[c=3]{c} \textbf{Masked Region Metrics} & & \\
    \textbf{Method} & XY & XZ & YZ & Mean & RMSE (HU) & MAE (HU) & SSIM ($\uparrow$) \\
    \midrule
    Gaussian Noise $\mathcal{N}(\mu, 200.0)$  & $19.3888$ & $16.8558$ & $16.5952$ & $17.6133$ & $519.6376$ & $424.0856$ & $0.0395$ \\
    Sample mean $\mu$ & $4.8369$ & $5.1570$ & $5.1722$ & $5.0554$ & $445.61$ & $379.27$ & $0.2935$ \\
    Lungs-DDPM & $15.0303$ & $13.9117$ & $14.1776$ & $14.3732$ & $439.53$ & $235.60$ & $0.6608$ \\
    MAISIv2 & $0.4369$ & $0.5297$ & $0.4996$ & $0.4887$ & $278.71$ & $147.25$ & $0.6777$ \\
    Our Method & $0.0439$ & $0.0400$ & $0.0410$ & $0.0417$ & $124.51$ & $65.80$ & $0.5972$ \\
    \bottomrule
    \end{tblr}
    \label{tab:baseline_comparison}
\end{table*}

\subsection{Expert validation} 

We performed two distinct experiments to evaluate the realism of the synthetic samples resulting from nodule removal and insertion. For this purpose, two expert radiologists evaluated the images using both consumer-grade and medical-grade displays, and domain-standard medical imaging software \citep{Kikinis2014}. Because our overall approach involves preserving the majority of the original image exactly, with modifications restricted to select cubic regions, each experiment presented the potentially modified region at the center of a cropped 3D CT scan. Every crop included a surrounding large enough offset to ensure that genuine local content served as a visual reference. This design ensures that experts do not need to search the entire scan for localized changes, allowing them to concentrate solely on the region of interest.

\subsection{\nshap attribution stability}

\begin{figure}[H]
    \centering
    \includegraphics[width=0.98\linewidth]{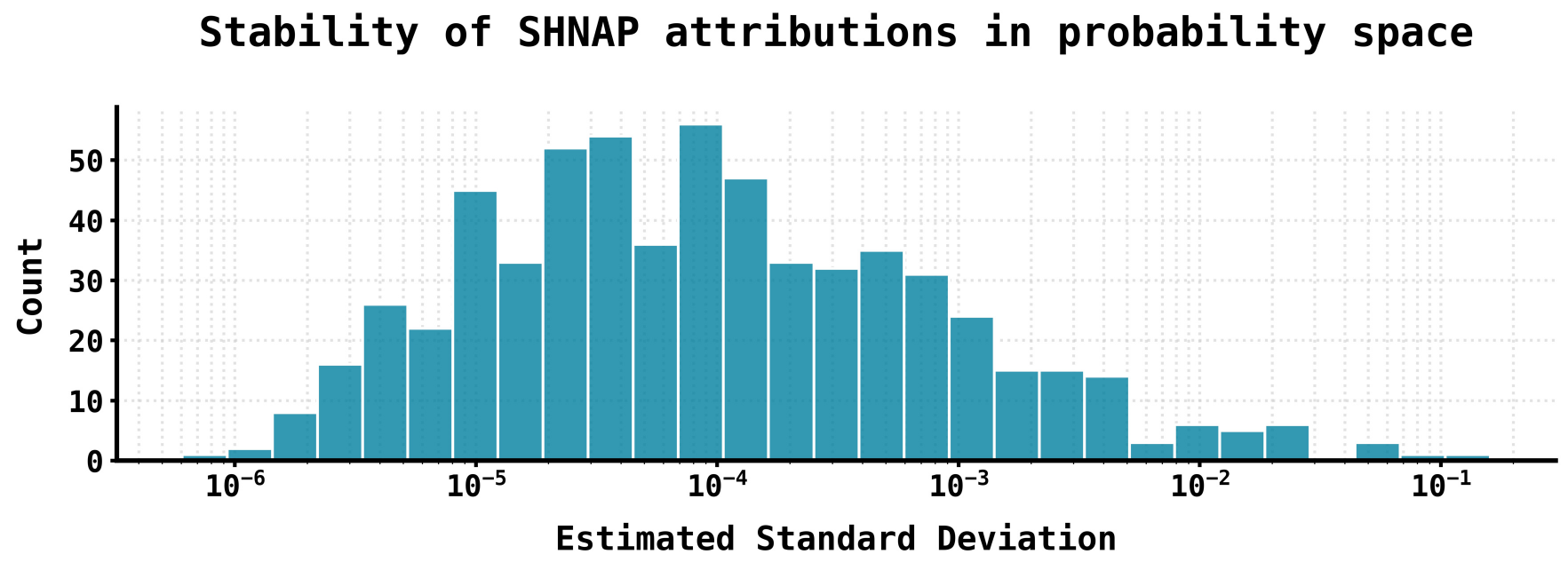}
    \caption{Histogram of standard deviations for all \nshap attributions from iLDCT. Attributions were computed in probability space to facilitate the interpretation of deviation magnitudes.}
    \label{fig:validity_stability}
\end{figure}

\begin{figure}[th]
    \centering
    \includegraphics[width=0.98\linewidth]{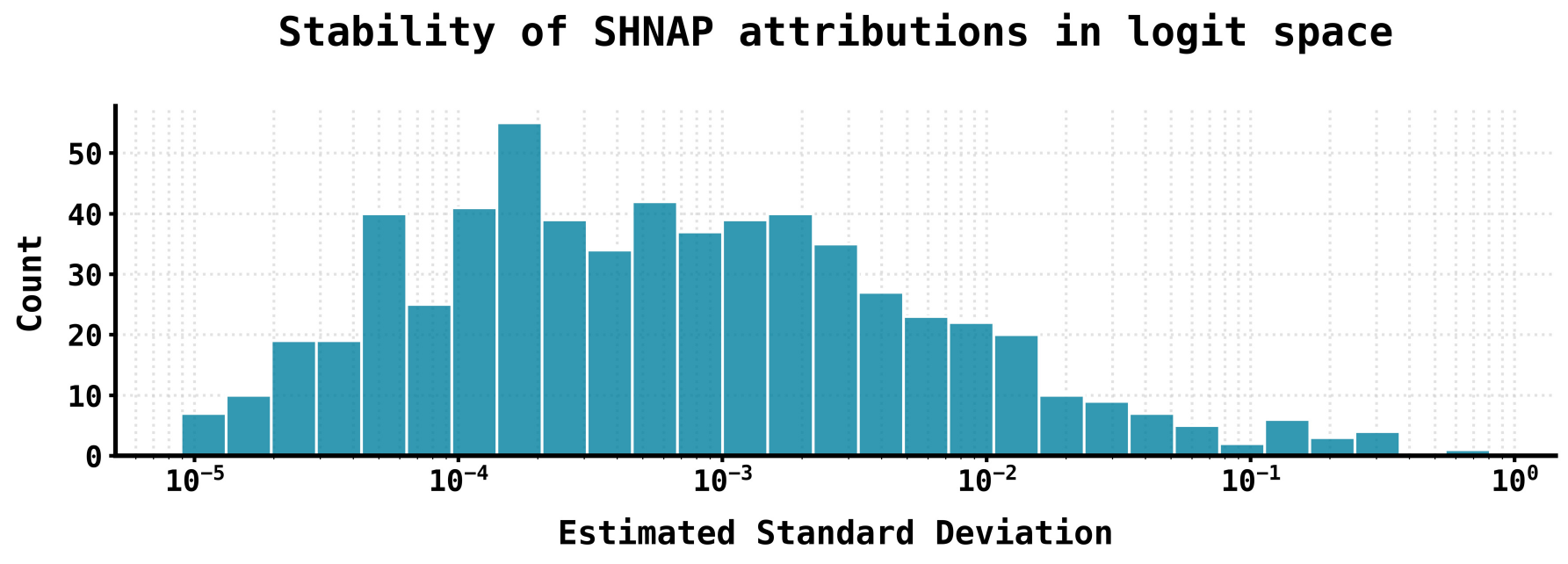}
    \caption{Histogram of standard deviations for all \nshap attributions from iLDCT. Attributions were computed in logit space.}
    \label{fig:validity_stability_logit}
\end{figure}

To assess the stability of the explanations, \Cref{fig:validity_stability} presents the distribution of standard deviations for all \nshap attributions across the iLDCT dataset. By computing these attributions in probability space, we provide a direct measure of local explanation variance. The tight clustering of low standard deviation values indicates that the \nshap estimates are highly stable, ensuring that the reported nodule contributions are not artifacts of sampling noise or local model instabilities. \Cref{fig:validity_stability_logit} presents an analogous plot in logit space.

\subsection{\nshap attribution stability across different baselines}\label{app:naive_stability}

\begin{figure}[th]
    \centering
    \includegraphics[width=0.98\linewidth]{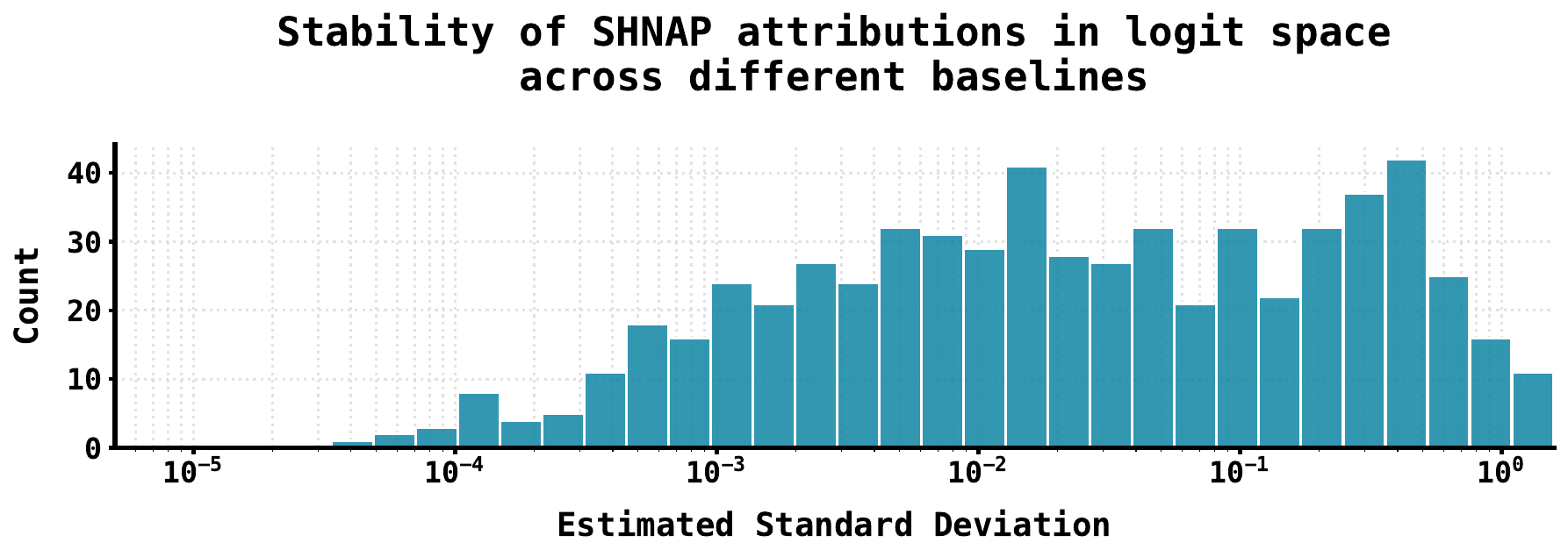}
    \caption{Histogram of standard deviations for all \nshap attributions from iLDCT on simple baselines. Attributions were computed in logit space.}
    \label{fig:validity_stability_logit_baseline}
\end{figure}

Selecting appropriate baselines for nodule removal is non-trivial. Simple strategies often yield out-of-distribution samples, leading to inconsistent model interpretations. We quantify this inconsistency in \Cref{fig:validity_stability_logit_baseline} by reporting the standard deviation of \nshap attributions computed across four distinct naive baselines for each image. Specifically, for every masked sample, we generated four separate variations using: global mean replacement, unmasked-region mean replacement, median replacement, and fixed lung tissue intensity. The resulting standard deviations of \nshap attributions are, on average, orders of magnitude higher than those observed with our method (\Cref{fig:validity_stability_logit}). This result indicates that naive baselines are inconsistent while our approach behaves stability and evaluates the model on the in-manifold data.

\subsection{Additional examples of \nshap}\label{app:shnap_extended}

\begin{figure}[ht]
    \centering
    \includegraphics[width=0.98\linewidth]{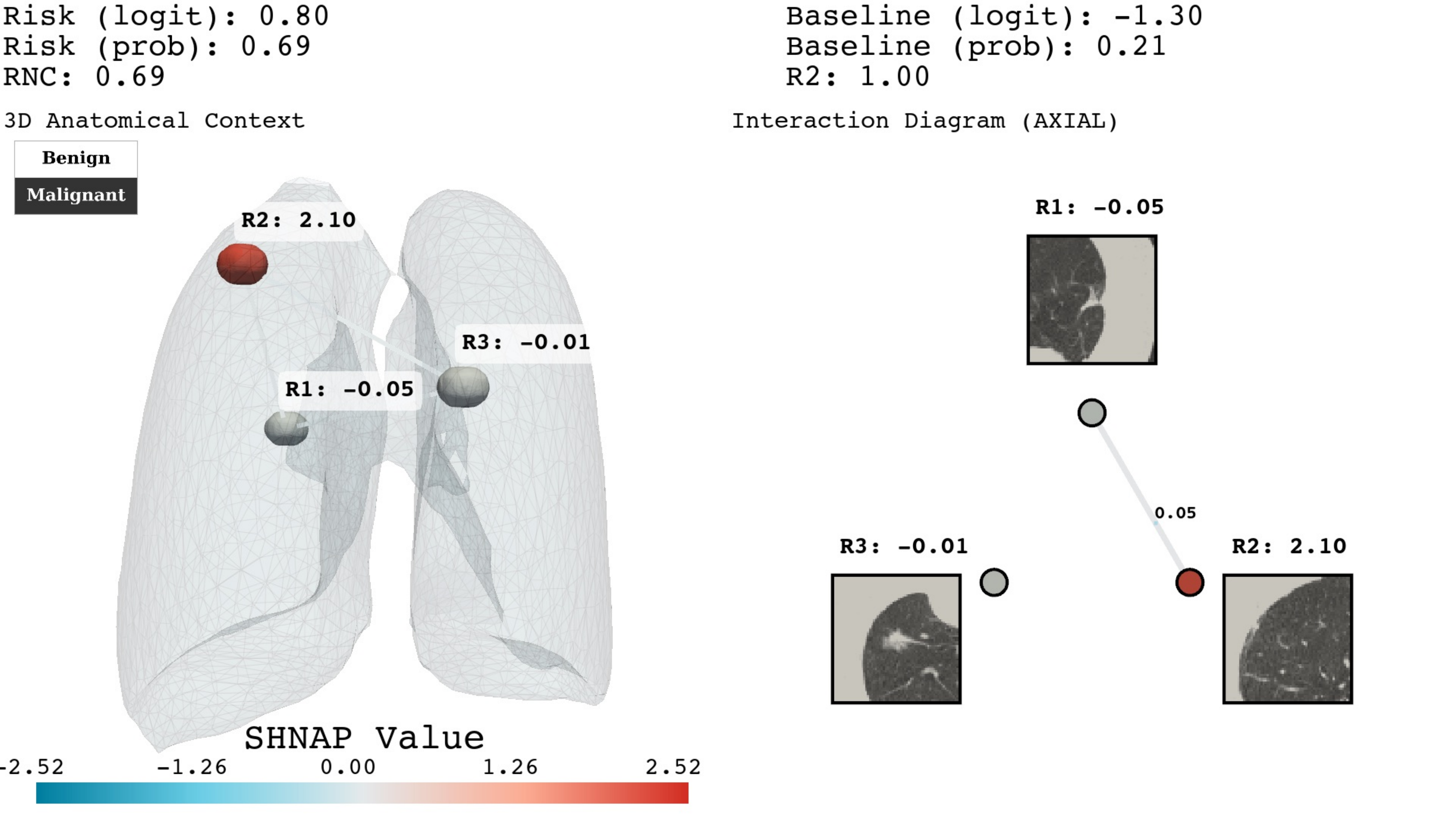}
    \caption{\nshap explanation visualizing 3D anatomical context and nodule interaction effects. This case demonstrates a follow-up scan of \cref{fig:nshap_nodules}~(\textbf{b}) featuring a false positive prediction where Sybil shifts focus to a different nodule than the one identified in \cref{fig:nshap_nodules}~(\textbf{b}).}
    \label{fig:shnap_extended_b_1}
\end{figure}

\begin{figure}[ht]
    \centering
    \includegraphics[width=0.98\linewidth]{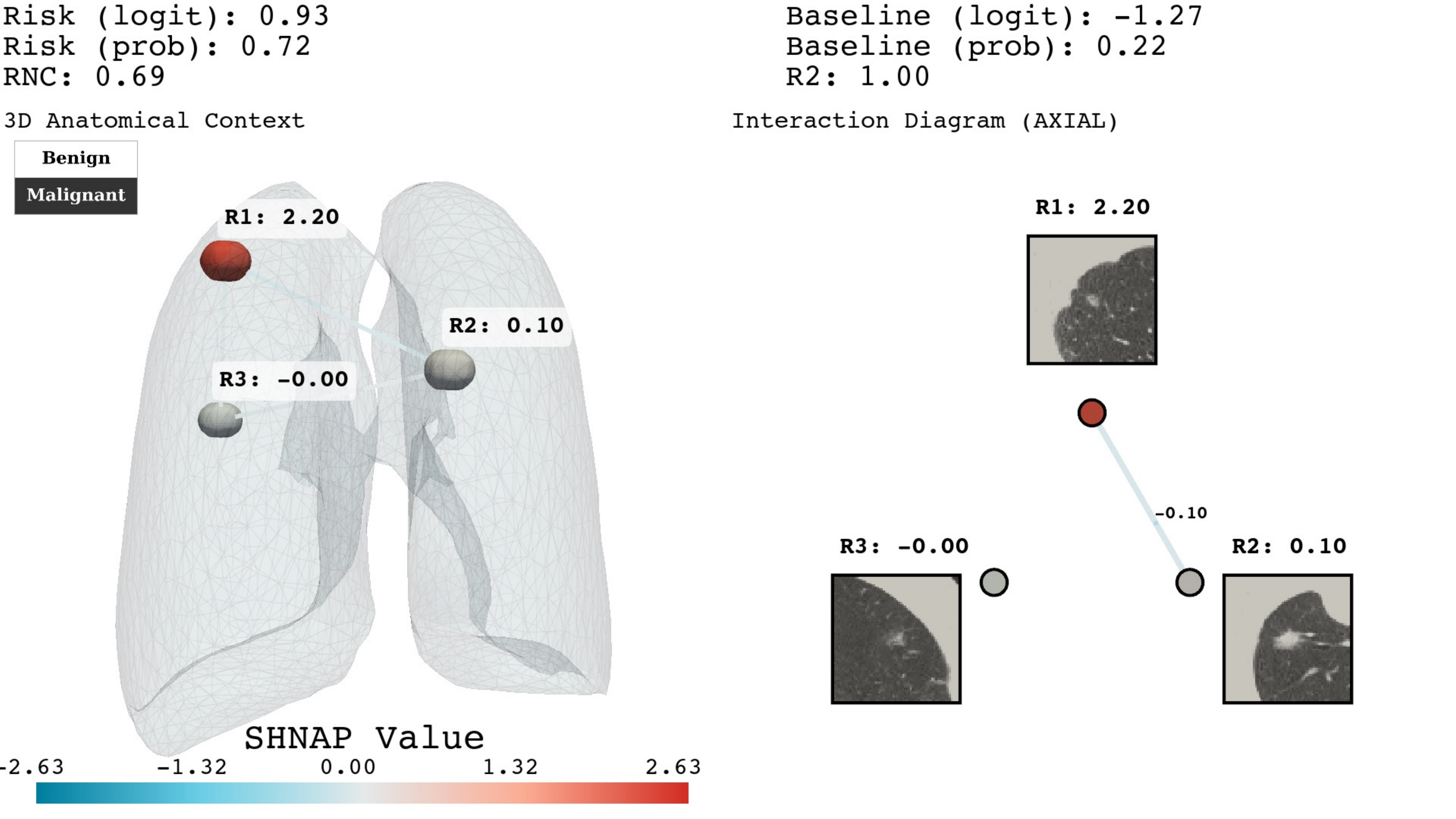}
    \caption{\nshap explanation visualizing 3D anatomical context and nodule interaction effects. This example extends \cref{fig:shnap_extended_b_1}, illustrating that Sybil does not retain its original focus from \cref{fig:nshap_nodules}~(\textbf{b}).}
    \label{fig:shnap_extended_b_2}
\end{figure}

\begin{figure}[ht]
    \centering
    \includegraphics[width=0.98\linewidth]{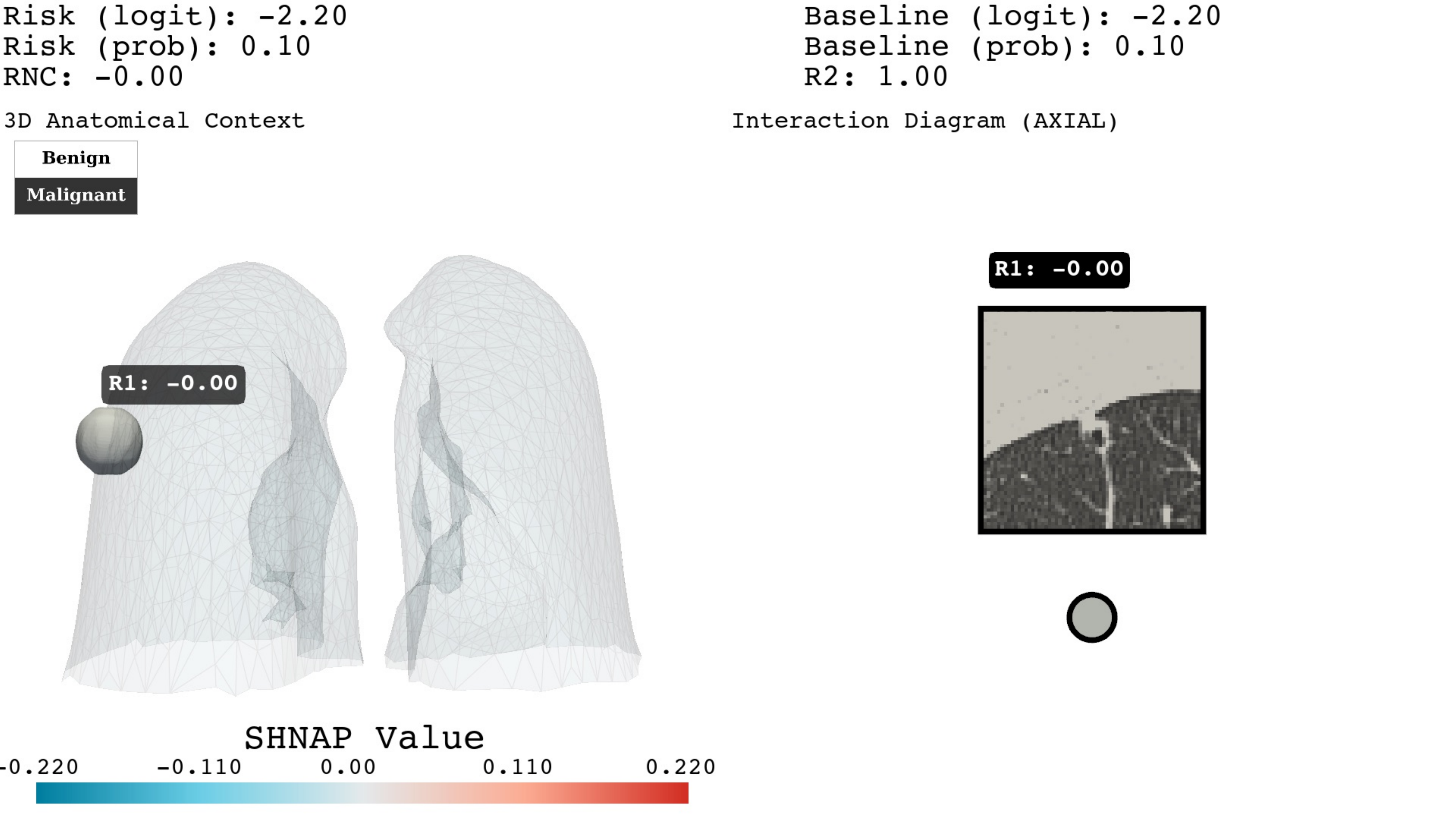}
    \caption{\nshap explanation visualizing 3D anatomical context and nodule interaction effects. The map reveals a pleural nodule receiving negligible attribution, which contributes to its omission from the model's risk assessment.}
    \label{fig:shnap_extended_c_1}
\end{figure}

\begin{figure}[ht]
    \centering
    \includegraphics[width=0.98\linewidth]{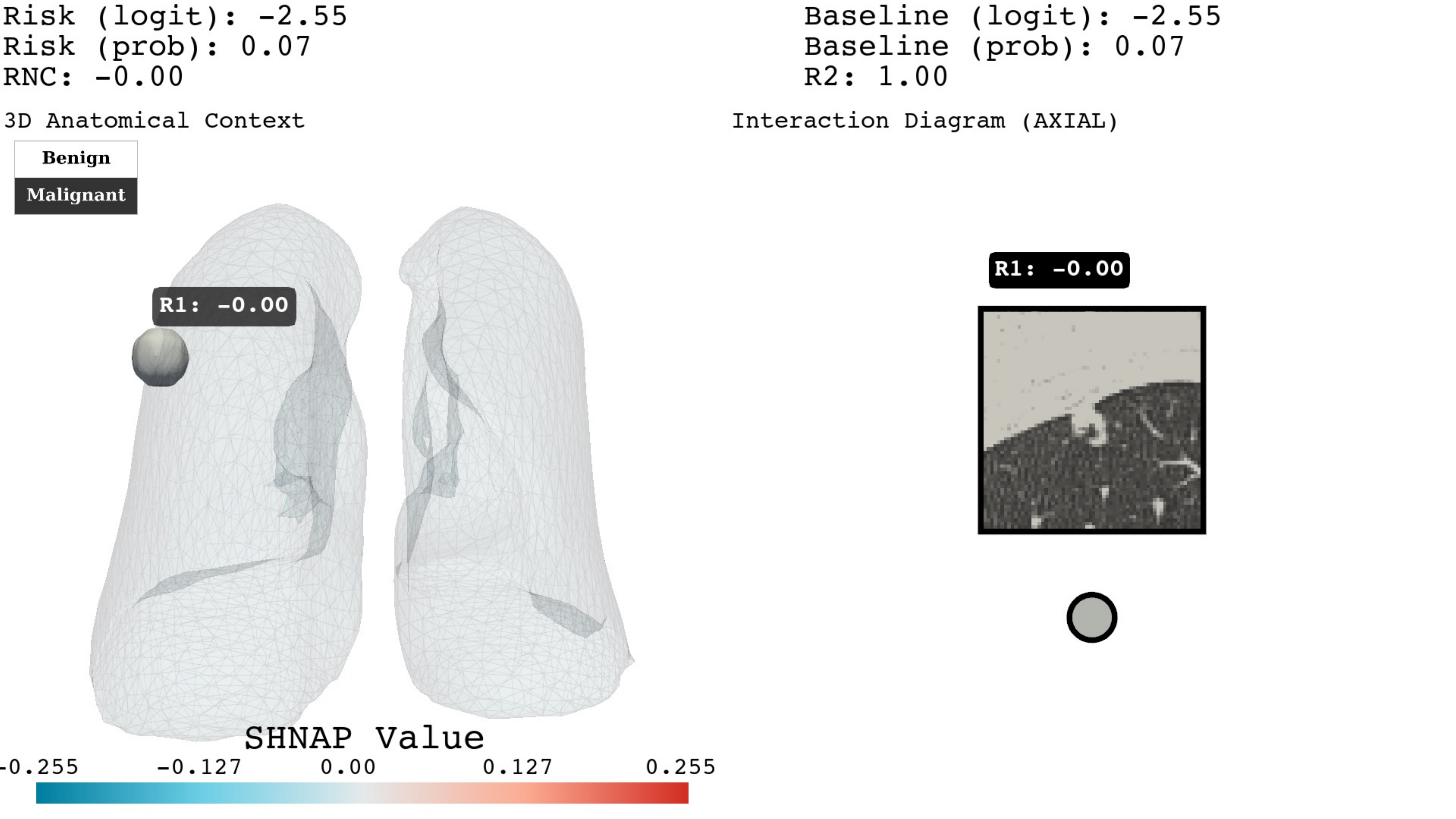}
    \caption{\nshap explanation visualizing 3D anatomical context and nodule interaction effects. Longitudinal follow-up of \cref{fig:shnap_extended_c_1} confirms a persistent failure to detect the peripheral pleural nodule across consecutive screening rounds.}
    \label{fig:shnap_extended_c_2}
\end{figure}

\begin{figure}[ht]
    \centering
    \includegraphics[width=0.98\linewidth]{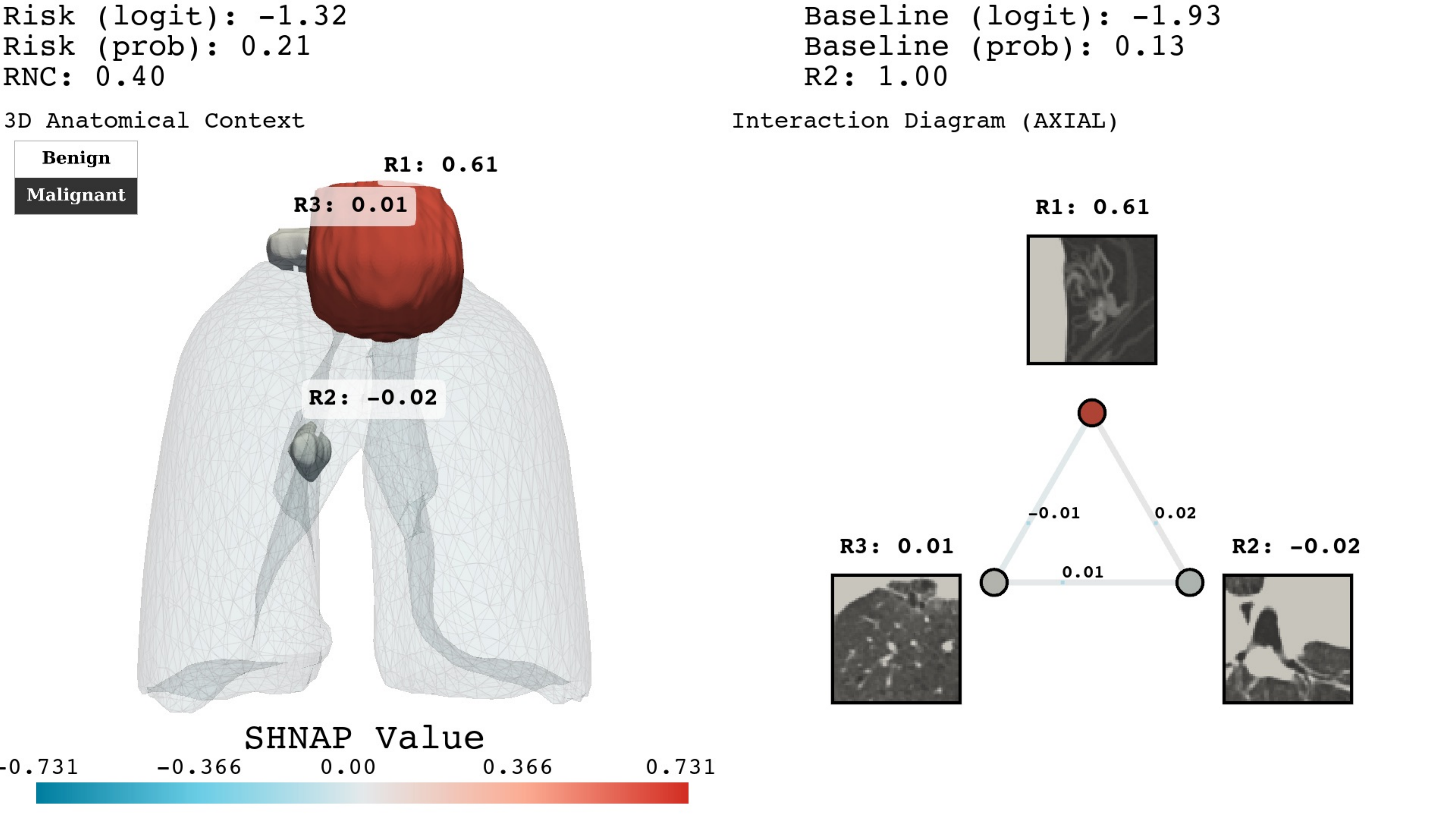}
    \caption{\nshap explanation visualizing 3D anatomical context and nodule interaction effects. The model erroneously relies on a thyroid goiter as a primary indicator of risk—a clinically unjustified diagnostic artifact.}
    \label{fig:shnap_extended_f_1}
\end{figure}

\begin{figure}[ht]
    \centering
    \includegraphics[width=0.98\linewidth]{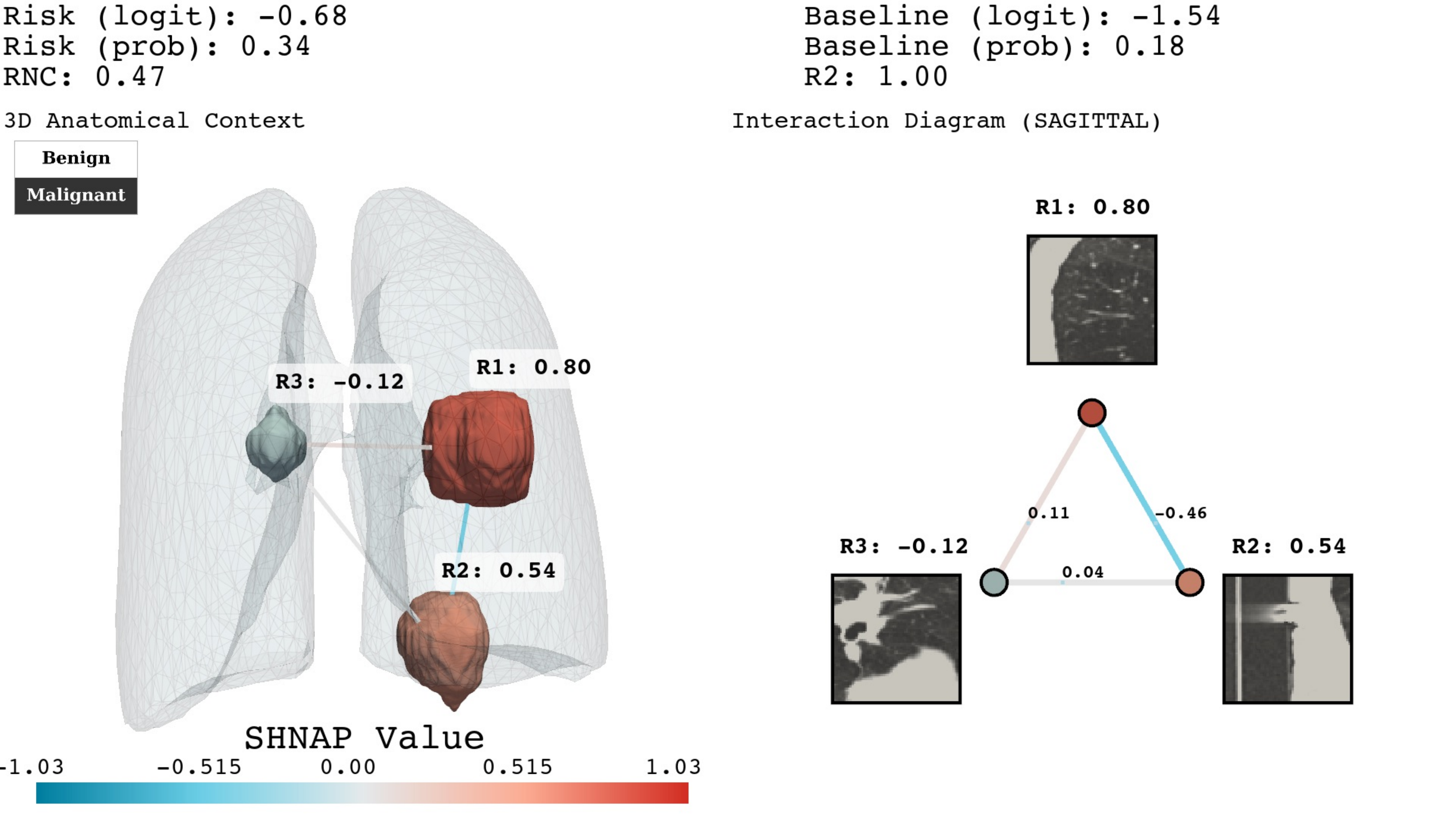}
    \caption{\nshap explanation visualizing 3D anatomical context and nodule interaction effects. Significant attribution is incorrectly placed on a metallic object positioned behind the patient's back, outside of the pulmonary region.}
    \label{fig:shnap_extended_f_2}
\end{figure}

We include additional \nshap explanations that extend the exploration from \cref{fig:nshap_nodules}. \Cref{fig:shnap_extended_b_1,fig:shnap_extended_b_2} show false positives characterized by multi-year instability. \Cref{fig:shnap_extended_c_1,fig:shnap_extended_c_2} visualize a pleural nodule missed by Sybil in year-after-year follow-up scans. \Cref{fig:shnap_extended_f_1} presents Sybil's reliance on a thyroid goiter, while \cref{fig:shnap_extended_f_2} illustrates how the model focuses on a metallic object behind the patient's back.

\subsection{Contribution of attention-based regions}\label{app:nshap_statistics_attention}

\begin{figure}[h]
    \centering
    \includegraphics[width=0.49\linewidth]{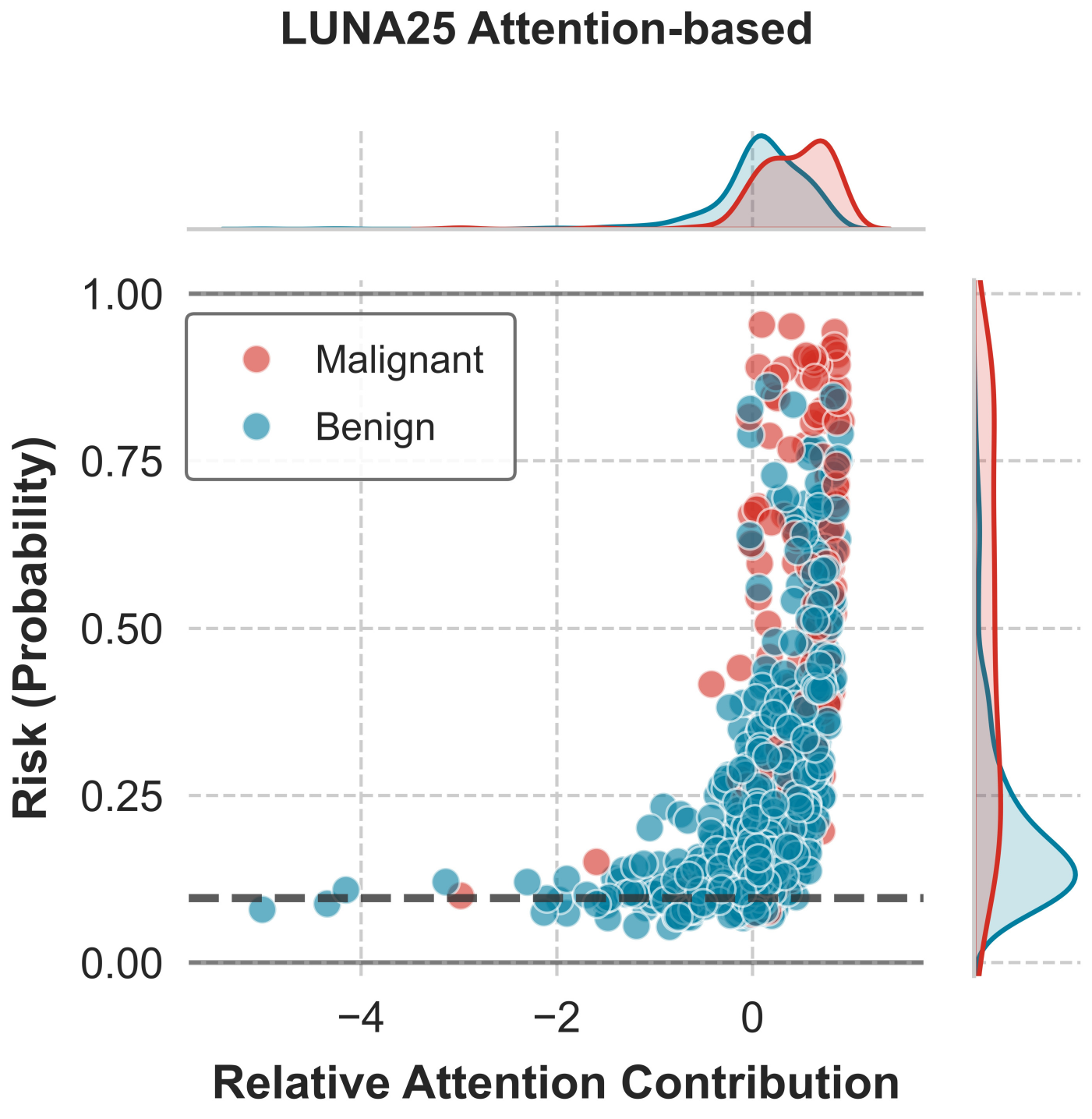}
    \includegraphics[width=0.49\linewidth]{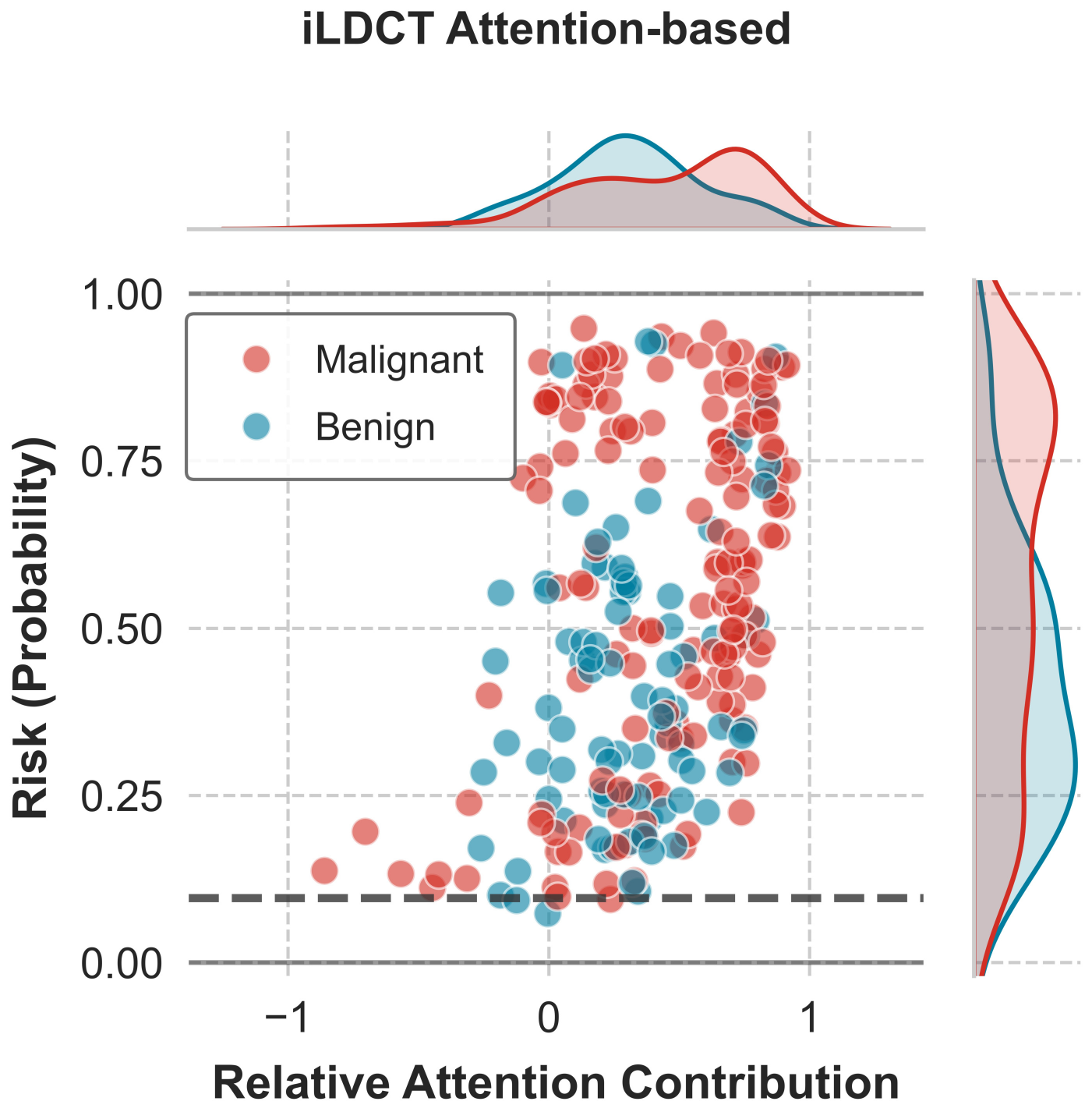}
    \caption{Comparison of Sybil's risk predictions against attention-based region contributions. The plots illustrate a shift in the importance distribution, highlighting a greater emphasis on the negative contributions within attention-based regions.}
    \label{fig:nshap_statistics_attention}
\end{figure}

\Cref{fig:nshap_statistics_attention} compares Sybil's base risk predictions against the relative contributions of attention-based regions, computed analogously to \her. A visible divergence in attribution patterns exists when compared to \cref{fig:nshap_statistics}, as the attention-based approach focuses more heavily on negative contributions.

\subsection{Influential regions are sparse}

\begin{figure}[h]
    \centering
    \includegraphics[width=0.98\linewidth]{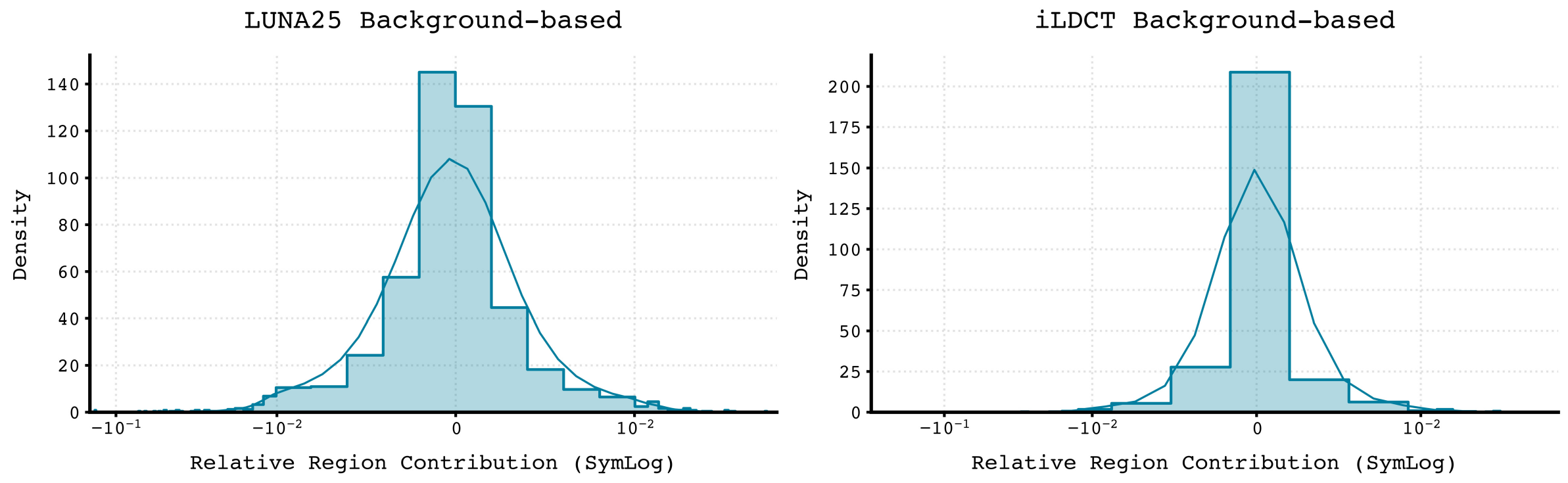}
    \caption{\nshap attribution for background lung regions. The distribution, plotted on a symlog scale, shows that regions excluding nodules and attention-based areas contribute negligibly to the overall risk prediction.}
    \label{fig:nshap_backgrounds}
\end{figure}

\Cref{fig:nshap_backgrounds} visualizes the relative contributions (on a symlog scale) of lung regions distinct from both annotated nodules and attention-based regions. These results indicate that the attribution for these background areas is effectively zero.

\subsection{\ninject attribution maps}\label{app:ninject_attribution_maps}

\begin{figure}[h]
    \centering
    \includegraphics[width=0.98\linewidth]{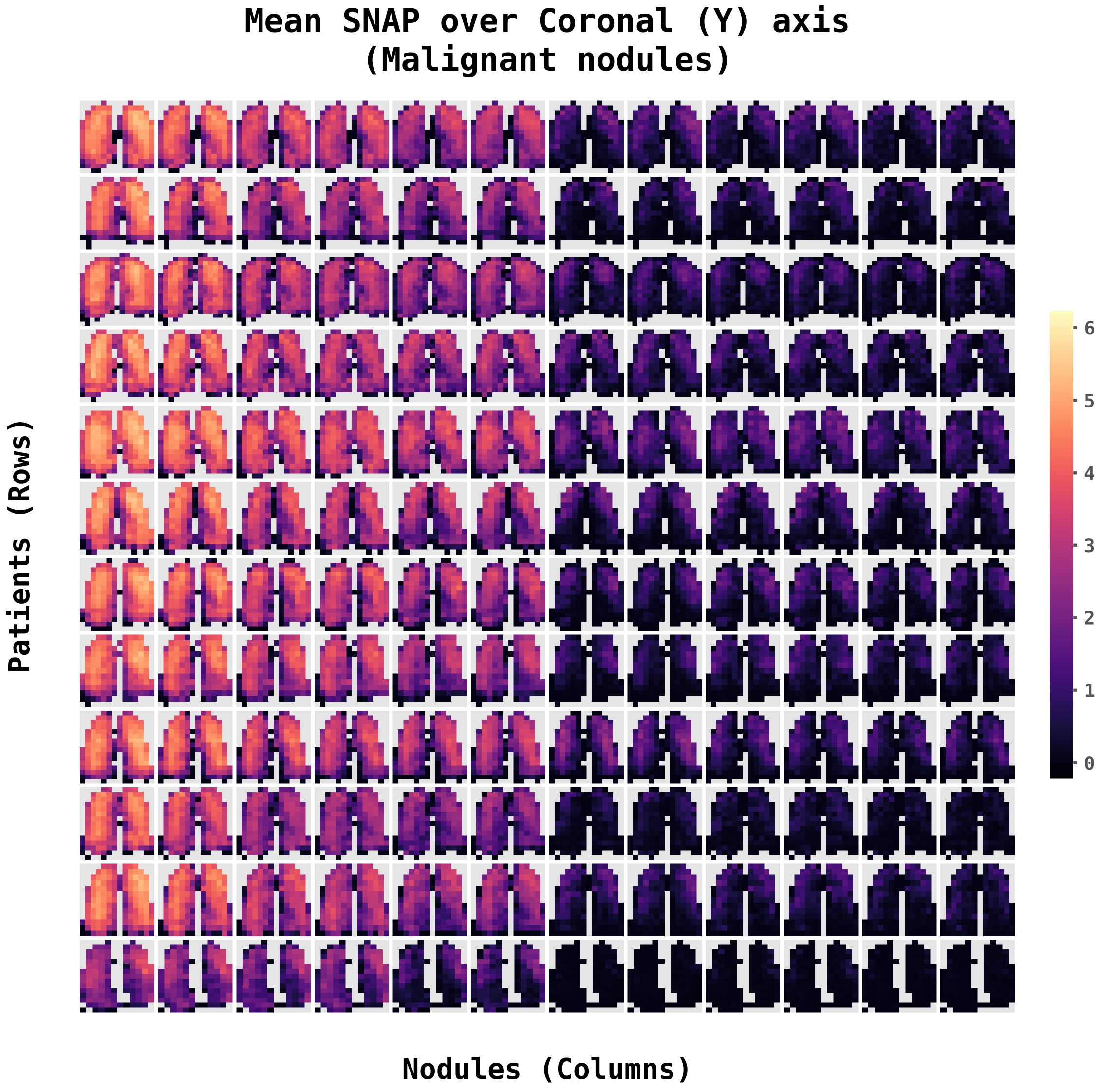}
    \caption{\ninject attribution maps for malignant nodules across patients in the coronal plane.}
    \label{fig:snap_attributions_a}
\end{figure}

\begin{figure}[h]
    \centering
    \includegraphics[width=0.98\linewidth]{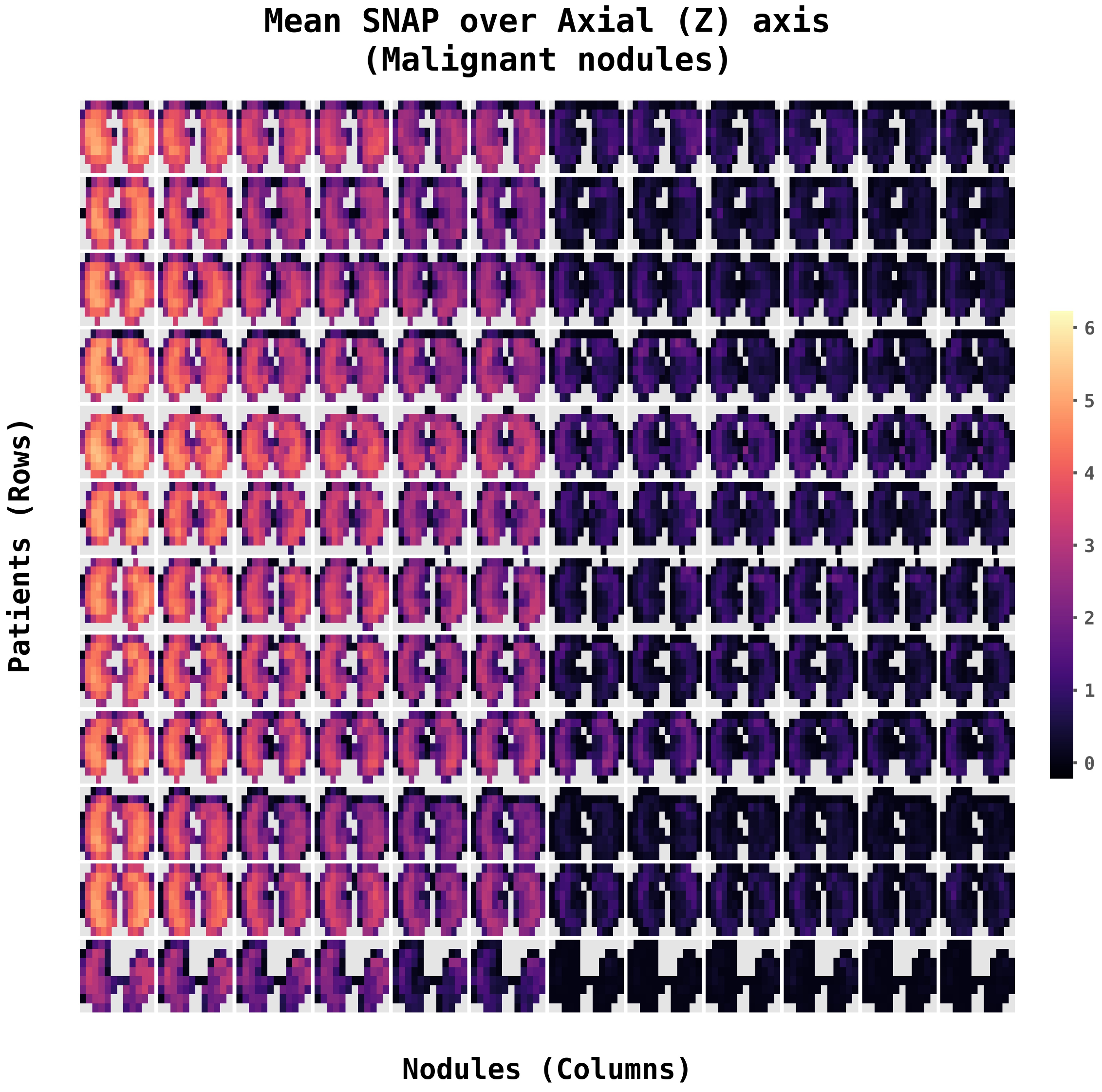}
    \caption{\ninject attribution maps for malignant nodules across patients in the axial plane.}
    \label{fig:snap_attributions_b}
\end{figure}

\begin{figure}[h]
    \centering
    \includegraphics[width=0.98\linewidth]{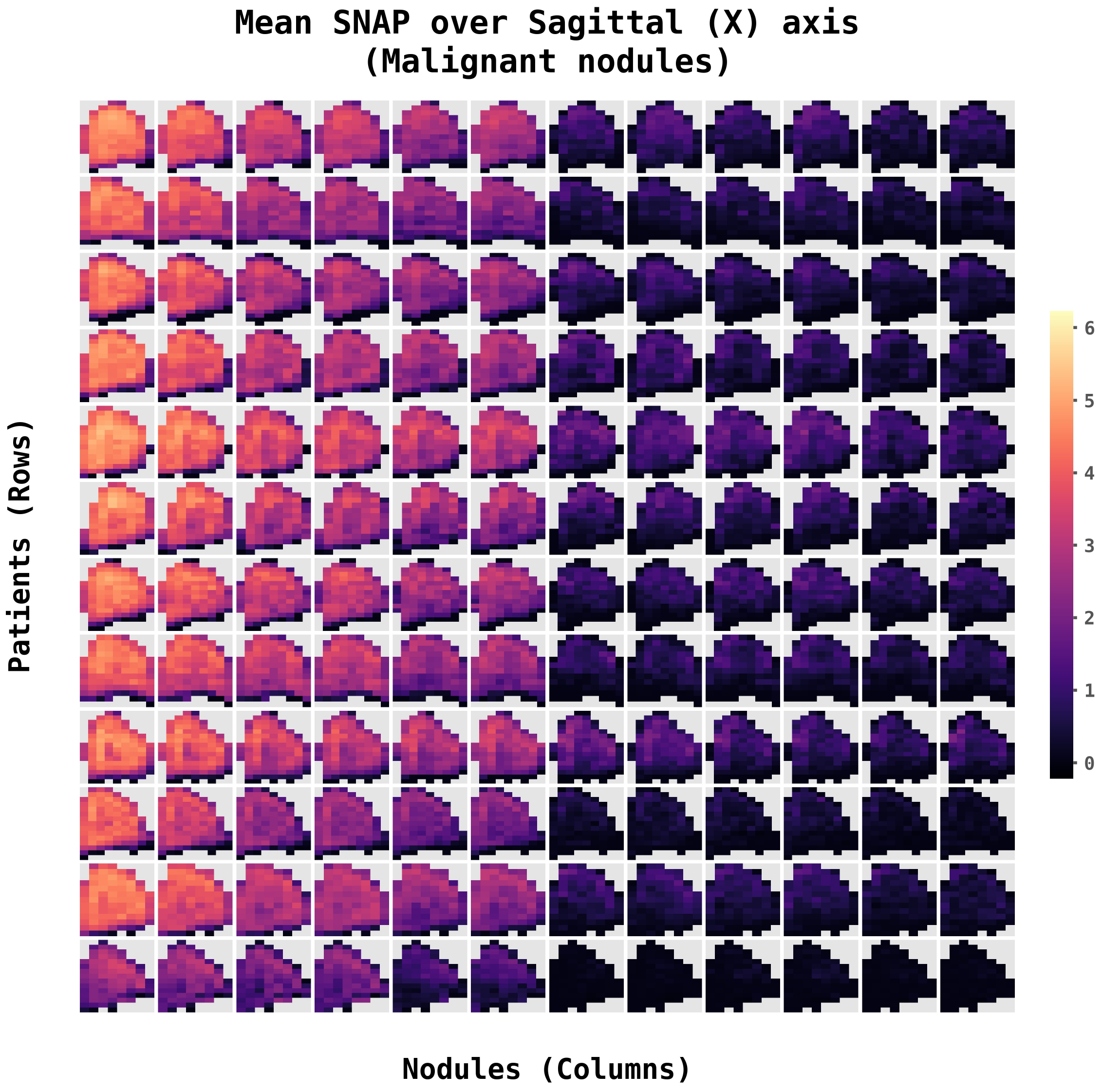}
    \caption{\ninject attribution maps for malignant nodules across patients in the sagittal plane.}
    \label{fig:snap_attributions_c}
\end{figure}

\Cref{fig:snap_attributions_a,fig:snap_attributions_b,fig:snap_attributions_c} depict \ninject attribution maps across 12 patient scans and 10 malignant nodules in coronal, axial, and sagittal planes, respectively. 

\begin{figure}[h]
    \centering
    \includegraphics[width=0.98\linewidth]{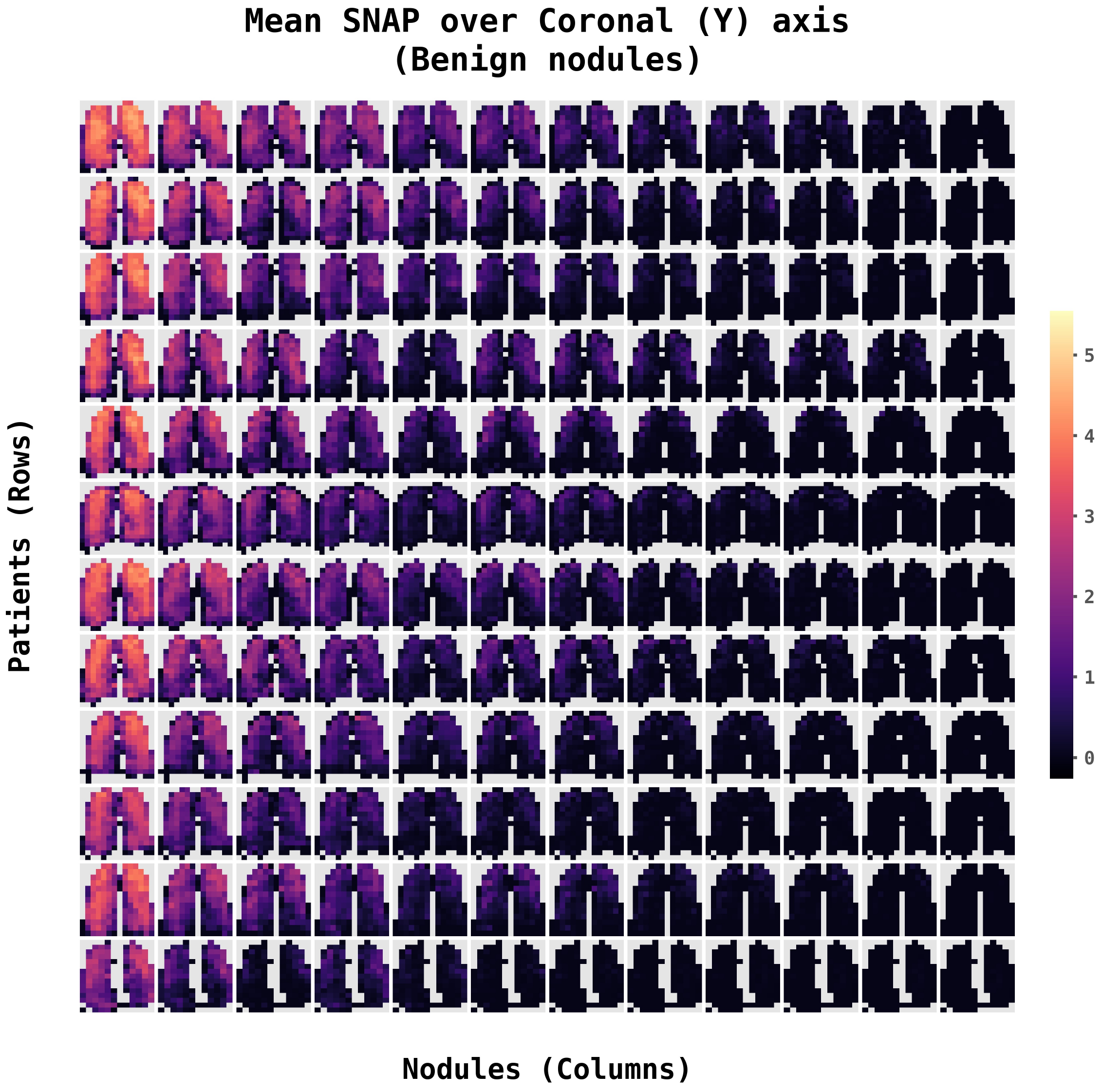}
    \caption{\ninject attribution maps for benign nodules across patients in the coronal plane.}
    \label{fig:snap_attributions_d}
\end{figure}

\begin{figure}[h]
    \centering
    \includegraphics[width=0.98\linewidth]{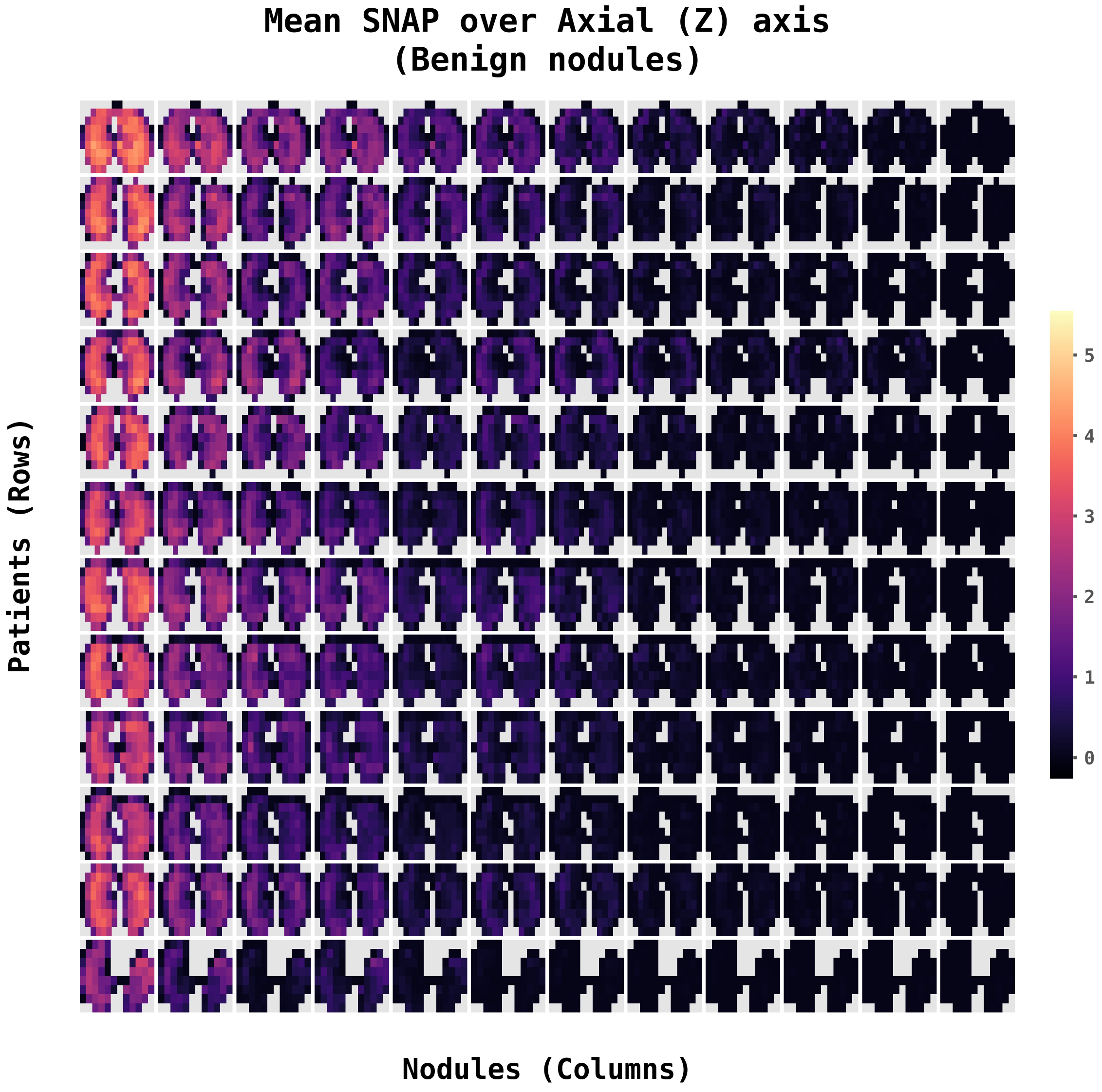}
    \caption{\ninject attribution maps for benign nodules across patients in the axial plane.}
    \label{fig:snap_attributions_e}
\end{figure}

\begin{figure}[h]
    \centering
    \includegraphics[width=0.98\linewidth]{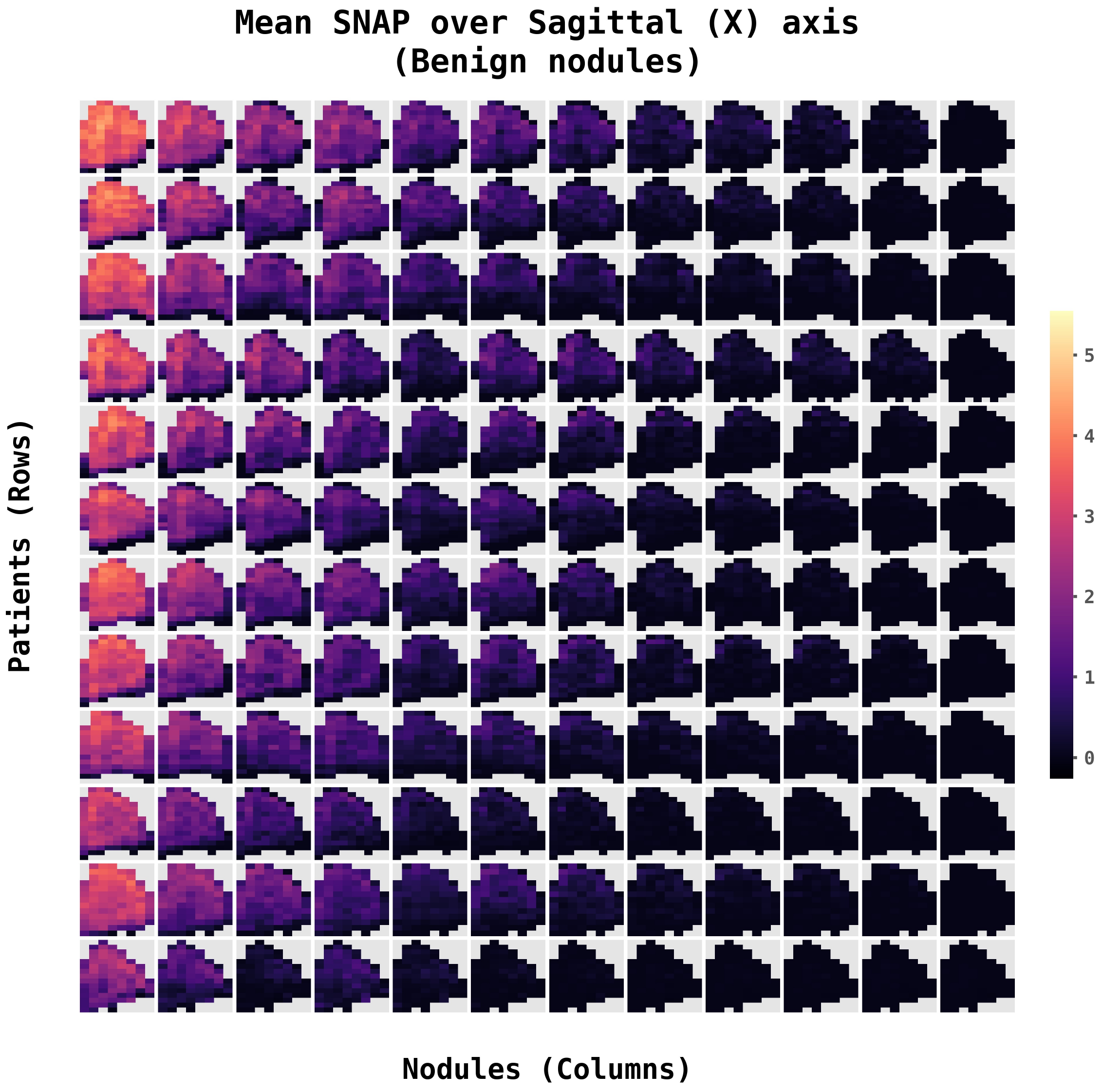}
    \caption{\ninject attribution maps for benign nodules across patients in the sagittal plane.}
    \label{fig:snap_attributions_f}
\end{figure}

\Cref{fig:snap_attributions_d,fig:snap_attributions_e,fig:snap_attributions_f} depict \ninject attribution maps across 12 patient scans and 10 benign nodules in coronal, axial, and sagittal planes, respectively.

\subsection{Radial sensitivity bias}

\begin{figure}[h]
    \centering
    \includegraphics[width=0.98\linewidth]{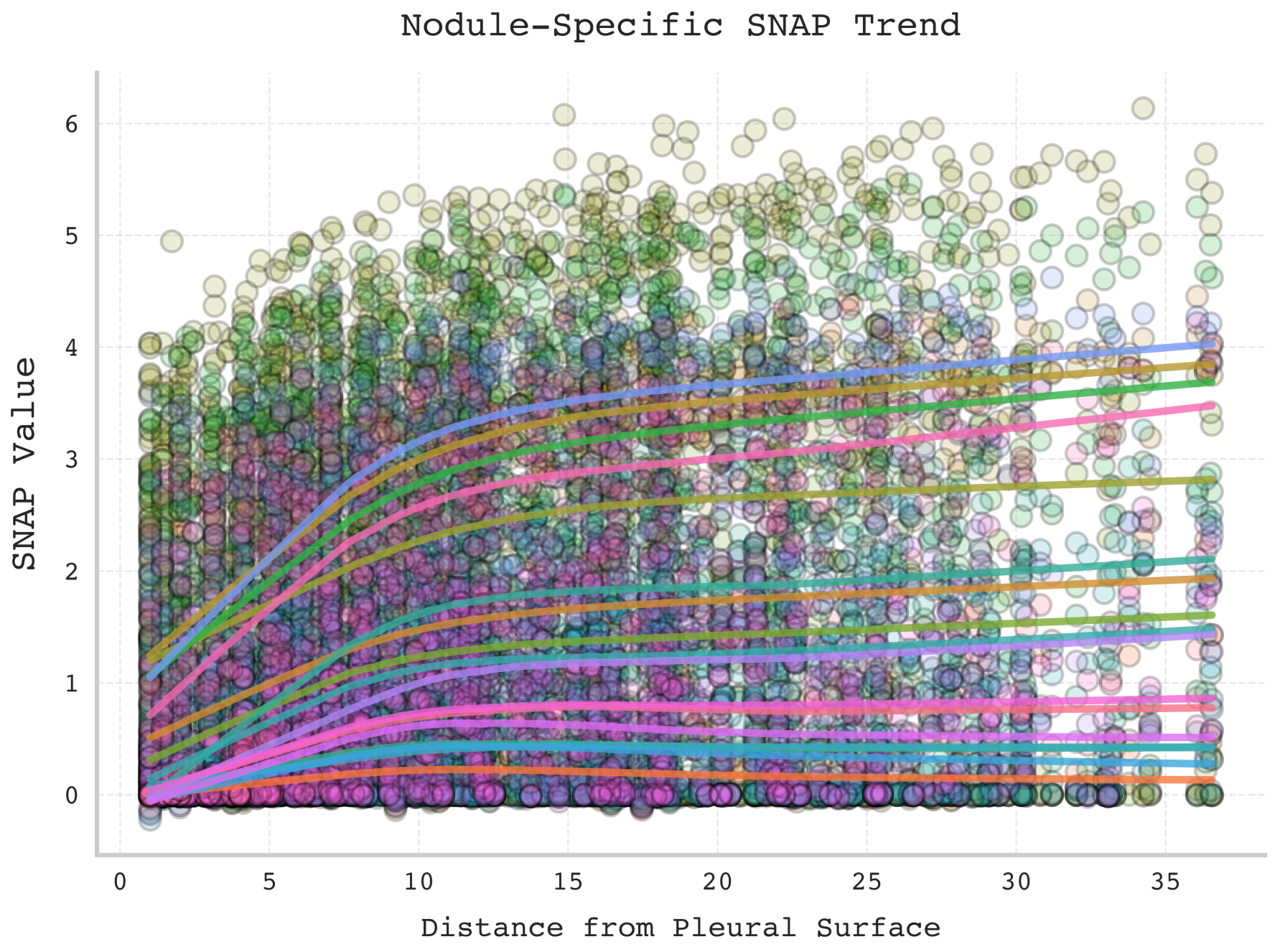}
    \caption{\ninject attribution values and corresponding trends across several nodules inserted into the same patient as a function of their distance from the pleural surface.}
    \label{fig:snap_radial_bias}
\end{figure}

\Cref{fig:snap_radial_bias} visualizes nonlinear trends for \ninject values across multiple nodules inserted into a single patient, plotted against each insertion's distance from the pleural surface.

\subsection{Qualitative examples}

\begin{figure}[h]
    \centering
    \includegraphics[width=0.98\linewidth]{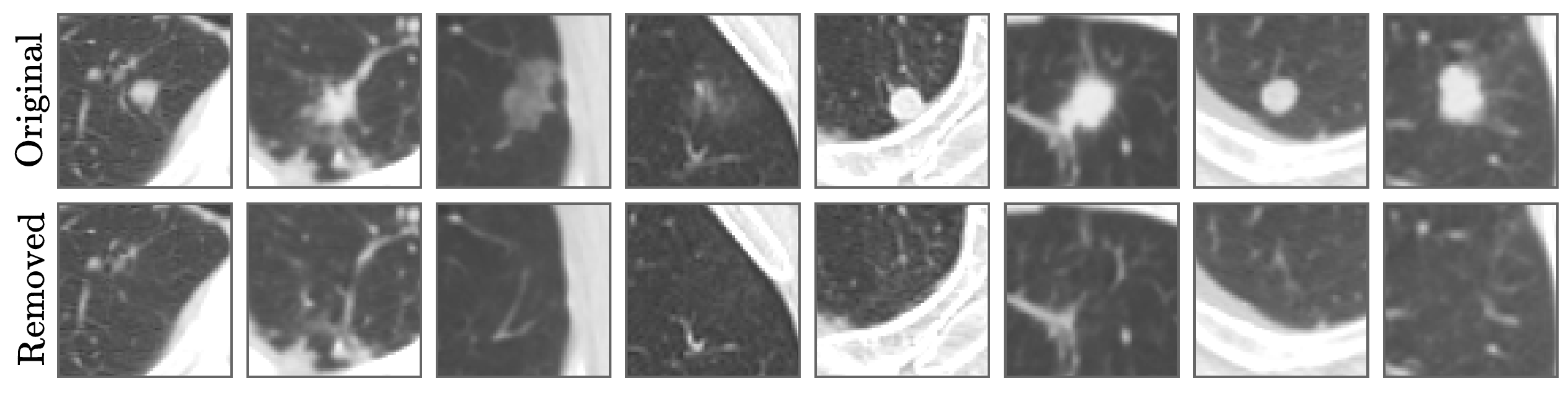}
    \caption{Qualitative examples of nodule removal.}
    \label{fig:nshap_examples}
\end{figure}

\begin{figure}[h]
    \centering
    \includegraphics[width=0.98\linewidth]{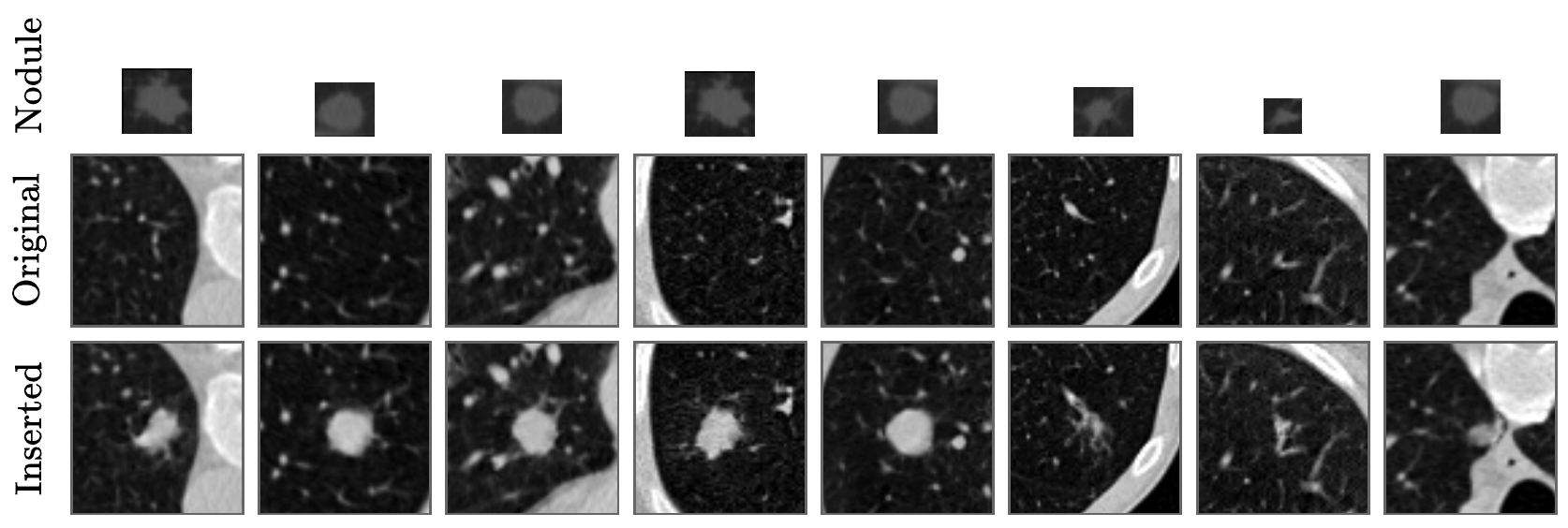}
    \caption{Qualitative examples of nodule insertion.}
    \label{fig:ninject_examples}
\end{figure}

We include qualitative examples of nodule removal and nodule insertion in \cref{fig:nshap_examples,fig:ninject_examples} respectively.

\end{document}